%% file: main.tex
\documentclass[preprint,12pt,times]{elsarticle}

\usepackage{amssymb}
\usepackage{amsmath}

\usepackage{lineno}

\usepackage{lipsum}
\usepackage{amsfonts}
\usepackage{graphicx}
\usepackage{epstopdf}
\usepackage{algorithm, algorithmic}

\usepackage{enumitem}
\setlist[enumerate]{leftmargin=.5in}
\setlist[itemize]{leftmargin=.5in}

\usepackage{amsopn}

\usepackage{caption}

\usepackage{hyperref}
\hypersetup{
    colorlinks=true,
    linkcolor=blue,
    filecolor=magenta,      
    urlcolor=cyan
}
\usepackage{multirow}
\usepackage{subcaption}
\usepackage[textsize=footnotesize, colorinlistoftodos]{todonotes}

\newcommand{\vx}{\mathbf{x}}

\newcommand{\vz}{\mathbf{z}}
\newcommand{\vt}{\mathbf{t}}
\newcommand{\vs}{\mathbf{s}}

\newcommand{\vu}{\mathbf{u}}
\newcommand{\vf}{\mathbf{f}}
\newcommand{\vmu}{\boldsymbol{\mu}}
\newcommand{\vSigma}{\boldsymbol{\Sigma}}

\newcommand{\vK}{\mathbf{K}}

\newcommand{\vA}{\mathbf{A}}
\newcommand{\vM}{\mathbf{M}}
\newcommand{\vb}{\mathbf{b}}
\newcommand{\vc}{\mathbf{c}}
\newcommand{\vzero}{\mathbf{0}}

\DeclareMathOperator*{\argmin}{arg\,min}

\journal{}
\date{}

\begin{document}

\begin{frontmatter}

\title{Fast Gaussian Processes under Monotonicity Constraints}

\author[label1]{Chao Zhang \corref{cor1}} 
\ead{chaz@dtu.dk}
\cortext[cor1]{Corresponding author}
\author[label1]{Jasper M. Everink} 
\ead{jmev@dtu.dk}
\author[label1]{Jakob Sauer Jørgensen}
\ead{jakj@dtu.dk}
  
\affiliation[label1]{organization={Department of Applied Mathematics and Computer Science, Technical University of Denmark},
            city={Kgs Lyngby},
            country={Denmark}}

\input{abstract}



\begin{keyword}
Gaussian processes, monotonicity constraints, inverse problems, Bayesian methods, constrained optimization



\end{keyword}

\end{frontmatter}
\thispagestyle{plain}

\input{introduction}
\input{gp}
\input{gp_under_constraints}
\input{experiments}
\input{applications}
\input{conclusions}

\section*{Data Availability}
The code used in this paper is available at \url{https://doi.org/10.5281/zenodo.15807484}.

\section*{Declaration of Competing Interest}
The authors declare that they have no known competing financial interests or personal relationships that could have appeared to influence the work reported in this paper.

\section*{Acknowledgments}
This work was funded by the Villum Foundation under Grant Number 25893.

\bibliographystyle{elsarticle-num}
\bibliography{references}


\end{document}

%% file: abstract.tex
\begin{abstract}
Gaussian processes (GPs) are widely used as surrogate models for complicated functions in scientific and engineering applications. In many cases, prior knowledge about the function to be approximated, such as monotonicity, is available and can be leveraged to improve model fidelity. Incorporating such constraints into GP models enhances predictive accuracy and reduces uncertainty, but remains a computationally challenging task for high-dimensional problems. In this work, we present a novel virtual point-based framework for building constrained GP models under monotonicity constraints, based on regularized linear randomize-then-optimize (RLRTO), which enables efficient sampling from a constrained posterior distribution by means of solving randomized optimization problems. We also enhance two existing virtual point-based approaches by replacing Gibbs sampling with the No U-Turn Sampler (NUTS) for improved efficiency. A Python implementation of these methods is provided and can be easily applied to a wide range of problems. This implementation is then used to validate the approaches on approximating a range of synthetic functions, demonstrating comparable predictive performance between all considered methods and significant improvements in computational efficiency with the two NUTS methods and especially with the RLRTO method. The framework is further applied to construct surrogate models for systems of differential equations.
\end{abstract}

%% file: introduction.tex
\section{Introduction}
\label{section:introduction}

Gaussian processes (GPs) are a versatile framework for function approximation, offering flexibility in modeling functions and inherent uncertainty quantification. Originating in geostatistics as \textit{kriging} \cite{chiles2018fifty}, GPs have become widely adopted in fields such as machine learning \cite{Williams2006gaussian, liu2020gaussian}, optimization \cite{frazier2018tutorial, morita2022applying}, and uncertainty quantification \cite{bilionis2013multi, chen2015uncertainty, Ernst2024uncertainty}. Their ability to build robust surrogate models makes them particularly valuable in scientific and engineering applications where data is scarce or costly. This paper addresses the computational challenge of incorporating monotonicity constraints into GP models, proposing faster methods to enhance their accuracy in the modeling of physical systems.

\subsection{Motivation}

In many applications, prior knowledge about the function to be approximated is available, often as physical constraints like boundedness, monotonicity, or convexity \cite{SwilerSurveyConstrained2020}. These constraints arise from fundamental principles; for example, chemical concentrations must remain non-negative, and temperatures are confined to monotonic relationships with respect to inputs like pressure or time. Incorporating such constraints into GP models improves predictive accuracy by aligning with known physical laws, reduces uncertainty by excluding infeasible solutions, and enhances data efficiency, which is critical in data-limited settings \cite{AgrellGaussianProcesses2019}.

Two primary approaches for enforcing monotonicity constraints in GPs are virtual point-based methods \cite{RiihimakiGaussianProcesses2010, DaVeigaGaussianProcess2012, AgrellGaussianProcesses2019, WangEstimatingShape2016, DaVeigaGaussianProcess2020} and spline-based methods \cite{maatouk2017gaussian, lopez2018finite, bachoc2022sequential, lopez2022high, zhou2024mass}. Virtual point-based methods \cite{RiihimakiGaussianProcesses2010} enforce constraints at a finite set of points, approximating global monotonicity. In contrast, spline-based methods \cite{maatouk2017gaussian} represent a sample from a GP as a linear combination of basis functions, translating monotonicity into constraints on the coefficients of the basis functions. Both approaches extend to other linear inequality constraints, such as boundedness and convexity. A recent review of techniques on constraining GPs is presented in \cite{SwilerSurveyConstrained2020}.

However, a shared challenge of both of the above mentioned approaches is their computational cost, as building constrained GPs boils down to sampling from a constrained posterior distribution, usually through Markov Chain Monte Carlo (MCMC) \cite{gamerman2006markov}, to estimate statistics, e.g., mean and variance. For virtual point-based methods, this involves sampling derivative values at virtual points, while spline-based methods require sampling basis function coefficients. The computational burden escalates in high-dimensional settings, where the number of virtual points or basis functions grows drastically. For example, the commonly used component-wise Gibbs sampler \cite{gelfand2000gibbs} in virtual point-based methods updates variables sequentially, leading to slow convergence when the number of virtual points is large or when variables are highly correlated.

To address the computational challenges of MCMC-based methods, the Regularized Linear Randomize-then-Optimize (RLRTO) framework has recently emerged as an efficient alternative for sampling from certain constrained posterior distributions with linear forward operators \cite{BardsleyMCMCMethod2012, bardsley2020mcmc, EverinkBayesianInference2023, everink2023sparse}. This framework transforms the task of posterior sampling into repeatedly solving constrained randomized linear least squares problems, which allows for using efficient optimization algorithms instead of samplers, and has been shown to be particularly effective in high-dimensional settings, where traditional MCMC-based sampling methods might struggle with computational efficiency.

\subsection{Our contributions}
This work advances constrained GP modeling with the following contributions:
\begin{itemize}
    \item We propose a novel virtual point-based method for enforcing monotonicity constraints in GPs by leveraging the RLRTO framework. This approach generates independent samples through efficient optimization algorithms. Unlike traditional MCMC methods, RLRTO eliminates the need for burn-in periods and rejection or sample thinning, as samples are inherently not sequentially correlated, significantly enhancing computational efficiency in high-dimensional settings. To our knowledge, this is the first application of RLRTO to enforce constraints in GPs.

    \item We demonstrate how existing virtual-point based GPs can be drastically accelerated by replacing the Gibbs-based sampling with the state-of-the-art No U-Turns Sampler (NUTS) \cite{hoffman2014no}. Making use of gradient information of a distribution, NUTS generally scales better with the dimension of the problem than the Gibbs sampler and, as far as the authors are aware, has not been applied to constraining GPs with virtual points before.
    \item Our implementation of constraining GPs with the RLRTO, Gibbs and NUTS approaches is publicly available at \url{https://doi.org/10.5281/zenodo.15807484}. The code allows for a range of different virtual point-based methods to be easily developed and can be applied to problems other than those considered in this work. It also allows reproducibility of our results.
    \item We demonstrate the flexibility of our proposed method on a variety of different problems in 1D, 2D and for applications in differential equations. We conduct extensive experiments comparing our RLRTO and NUTS methods against existing virtual point-based methods using Gibbs samplers. Both methods but especially the RLRTO approach achieves superior computational efficiency while maintaining comparable accuracy.
\end{itemize}

While our focus is on monotonicity constraints for clarity, our framework can extend to other linear inequality constraints, e.g., boundedness and convexity.

\subsection{Related Work}
In constrained Bayesian inference, posterior distributions are restricted to constrained supports. Gradient-free methods for sampling constrained posterior distributions include rejection sampling \cite{rao2016data}, Metropolis-Hastings \cite{chib1995understanding}, and Gibbs sampling \cite{kotecha1999gibbs}, while gradient-based methods include the Mirror Langevin Algorithm (MLA) \cite{zhang2020wasserstein},  Moreau-Yoshida Unadjusted Langevin Algorithm (MYULA) \cite{durmus2018efficient} and Hamiltonian Monte Carlo (HMC) \cite{pakman2014exact}. When the forward operator is linear, the RLRTO framework offers superior efficiency in high-dimensional settings, which we will leverage for constraining GPs.

Neural networks offer an alternative approach to enforcing monotonicity constraints \cite[e.g.][]{archer1993application, wehenkel2019unconstrained, nolte2023expressive, runje2023constrained, kitouni2023robust}. However, their lack of inherent uncertainty quantification makes GPs more suitable for applications requiring probabilistic predictions, such as physical systems modeling.

\subsection{Organization of this paper}
In the following, we first review fundamentals of GPs in Section~\ref{section:gp}. Section~\ref{section:gp-under-constraints} describes two existing virtual point-based methods for incorporating monotonicity constraints in GPs and corresponding samplers, followed by a discussion on our newly proposed method based on RLRTO. Section~\ref{section:experiments} presents a series of numerical experiments to compare our method with existing methods, including a comparison in both quality and computational cost. Section~\ref{section:applications} discusses the application of our method to more practical examples, particularly in the context of building surrogate models for systems of differential equations. Finally, we conclude the paper in Section~\ref{section:conclusion}.

\subsection{Notations}
Throughout this paper, we use the following notations:
\begin{itemize}
    \item Lowercase letters like $t$ denote scalars.
    \item Lowercase bold letters like $\vx$ denote vectors and $\vx(\vt)$ denotes a vector evaluated at $\vt$.
    \item Uppercase bold letters like $\vK$ denote matrices and $\vK(\vt, \vs)$ denotes a covariance matrix between the points $\vt$ and the points $\vs$.
\end{itemize}

%% file: gp.tex
\section{Gaussian Processes}
\label{section:gp}

\subsection{Derivative-free Gaussian Processes}
\label{section:vanilla-gp}

A GP is a type of stochastic process that generalizes the notion of a \textit{multivariate Gaussian distribution} to an infinite-dimensional setting \cite{Williams2006gaussian}. Specifically, it is a collection of random variables indexed by a set $\mathcal{T}$, which is often a subset of $\mathbb{R}^d$, such that any finite collection of these random variables follows a multivariate Gaussian distribution. Formally, let $\mathcal{T}\subset \mathbb{R}^d$ be a set of inputs. A stochastic process $f:\mathcal{T}\to\mathbb{R}$ is called a GP if for any finite set of points $\vt = \{t_1, t_2, \ldots, t_N\} \subset \mathcal{T}$, the random vector $\vf(\vt) = [f(t_1), f(t_2), \ldots, f(t_N)]^\top$ follows a multivariate Gaussian distribution. That is, 
\begin{equation}
    \vf(\vt) \sim \mathcal{N}\left(\vmu(\vt), \vK(\vt, \vt)  \right),\label{eq:gp}
\end{equation}
where $\vmu(\vt) = [\mu(t_1), \mu(t_2), \ldots, \mu(t_N)]^\top$ is the mean vector with $\mu:\mathcal{T}\to\mathbb{R}$ being the mean function and $\vK(\vt, \vt)$ is the covariance matrix with entries $[\vK(\vt, \vt)]_{ij} = k(t_i, t_j)$ and $k:\mathcal{T}\times\mathcal{T}\to\mathbb{R}$ being the covariance function, also known as the kernel function. We omit observation noise in this work for clarity, though it can be readily incorporated.

The covariance function characterizes the correlation structure of the random variables within the process and incorporates desired properties such as smoothness and periodicity. Commonly used covariance functions include the squared exponential, Matérn, and periodic kernels. In this work, we use the squared exponential kernel for the convenience that it is infinitely differentiable \cite{adler2010geometry}. The squared exponential kernel has the simple form:
\begin{equation}
    k(t, t^\prime) := \sigma^2\exp\left(-\frac{(t-t^\prime)^2}{2l^2}\right), \quad \text{for $t, t^\prime\in \mathcal{T}$}, \label{eq:squared-exponential}
\end{equation}
where the hyperparameter $\sigma^2
$ represents the variance and $l$ represents the correlation length.

Given observed function values $\vf(\vt)$ at a set of training points $\vt$, GP can be used to predict function values at a new set of points. The prediction is based on the fact that the joint distribution of the function values at the training and prediction points follows a multivariate Gaussian distribution:
\begin{equation}
    \begin{bmatrix}
        \vf(\vt)\\
        \vf(\vu)
    \end{bmatrix} \sim \mathcal{N}\left(
        \begin{bmatrix}
            \vmu(\vt)\\
            \vmu(\vu)
        \end{bmatrix}
        ,
        \begin{bmatrix}
            \vK(\vt, \vt) & \vK(\vt, \vu)\\
            \vK(\vu, \vt) & \vK(\vu, \vu)
        \end{bmatrix}
    \right),\label{eq:gp-joint}
\end{equation}
where $\vf(\vu)$ are the function values at prediction points $\vu$. $\vK(\cdot,\cdot)$ are covariance matrices by evaluating the covariance function, e.g., \eqref{eq:squared-exponential}, at corresponding points. The posterior distribution of the function values $\vf(\vu)$ given $\vf(\vt)$ is then Gaussian:
\begin{equation}
    \vf(\vu) | \vf(\vt) \sim \mathcal{N}(\vmu^*, \vSigma^*),
\end{equation}
where the posterior mean and covariance are given by
\begin{align}
    \vmu^* &= \vmu(\vu) + \vK(\vu, \vt)\vK^{-1}(\vt, \vt)(\vf(\vt)-\vmu(\vt)),\\
    \vSigma^* &= \vK(\vu, \vu) - \vK(\vu, \vt)\vK^{-1}(\vt, \vt)\vK(\vt, \vu).
\end{align}
We refer to this GP model as the \textit{derivative-free} GP since it does not utilize any derivative information of the function.

The hyperparameters within the covariance function $k(t,t^\prime)$ are generally determined by maximizing the log marginal likelihood of the training data:
\begin{equation}
    \begin{split}
    \log p(\vf(\vt) | \vt) = &-\frac{1}{2}(\vf(\vt) - \vmu(\vt))^\top\vK^{-1}(\vt, \vt)(\vf(\vt) - \vmu(\vt))\\
    &- \frac{1}{2}\log\left|\vK(\vt, \vt)\right| - \frac{N_{\vt}}{2}\log(2\pi),\label{eq:log-marginal-likelihood}
    \end{split}
\end{equation}
where $N_{\vt}$ is the number of training points in $\vt$. The optimization can be done using gradient-based optimization methods, e.g., L-BFGS-B.

\subsection{Derivative-enhanced Gaussian Processes}\label{subsec:enhanced}
\label{subsection:derivative-enhanced-gp}

It is well known that Gaussian distributions are closed under linear operations and any linear combination of Gaussian random variables is still Gaussian. Similarly, for Gaussian processes, applying differentiation, which is a linear operator, preserves the GP structure \cite{parzen1999stochastic}. Specifically, let $\vf^\prime(\vs)$ be the derivative values of the function at $\vs$, in addition to $\vf(\vt)$ being function values at $\vt$, then the joint distribution of the function values $\vf^\prime(\vs)$ and $\vf(\vt)$ is Gaussian:
\begin{equation}
    \begin{bmatrix}
        \vf(\vt)\\
        \vf^\prime(\vs)
    \end{bmatrix} \sim \mathcal{N}\left(
        \begin{bmatrix}
            \vmu(\vt)\\
            \vmu^\prime(\vs)
        \end{bmatrix}
        ,
        \begin{bmatrix}
            \vK(\vt, \vt) & \vK^{01}(\vt, \vs)\\
            \vK^{10}(\vs, \vt) & \vK^{11}(\vs, \vs)
        \end{bmatrix}
    \right),\label{eq:gp-joint-derivative}
\end{equation}
where $\vmu^\prime(\vs)$ is the derivative of the mean function $\mu$ at $\vs$. When $\mu(t)$ is chosen to be a constant function, which is the case in this study, $\vmu^\prime(\vs)$ will be a vector of zeros. $\vK^{01}$ and $\vK^{10}$ are the cross-covariance matrices of the function values and derivatives by evaluating the following cross-covariance functions at corresponding points:
\begin{equation}
    k^{01}(t, t^\prime) = \frac{\partial}{\partial t^\prime}k(t, t^\prime),\quad k^{10}(t, t^\prime) = \frac{\partial}{\partial t}k(t, t^\prime).
\end{equation}
$\vK^{11}$ is the covariance matrix of the derivative values by evaluating the following covariance function:
\begin{equation}
    k^{11}(t, t^\prime) = \frac{\partial^2}{\partial t\partial t^\prime}k(t, t^\prime).
\end{equation}
In the case of the squared exponential covariance function in \eqref{eq:squared-exponential} and $d=1$, the derivatives of the covariance function are given by
\begin{equation}
    \begin{aligned}
        k^{01}(t, t^\prime) &= \sigma^2\exp\left(-\frac{(t-t^\prime)^2}{2l^2}\right)\frac{t-t^\prime}{l^2},\\
        k^{10}(t, t^\prime) &= \sigma^2\exp\left(-\frac{(t-t^\prime)^2}{2l^2}\right)\frac{t-t^\prime}{l^2}, \text{ and}\\
        k^{11}(t, t^\prime) &= \sigma^2\exp\left(-\frac{(t-t^\prime)^2}{2l^2}\right)\left(\frac{(t-t^\prime)^2}{l^4}-\frac{1}{l^2}\right).
    \end{aligned}
\end{equation}

Based on \eqref{eq:gp-joint-derivative}, we are able to make predictions of derivatives based on function values, and vice versa. For example, given the derivative values $\vf^\prime(\vs)$, we can estimate $\vf(\vt)$ using the following conditional mean:
\begin{equation}
    \vmu(\vt) + \vK^{01}(\vt, \vs)\vK^{11}(\vs, \vs)^{-1}(\vf^\prime(\vs)-\vmu^\prime(\vs)), \label{eq:gp-predict-value-based-on-derivative}
\end{equation}
which defines an affine mapping from the derivative values $\vf^\prime(\vs)$ to the function values $\vf(\vt)$, and this estimation will be heavily used in Section~\ref{section:gp-under-constraints}.
The uncertainty of this estimation is quantified by the conditional covariance:
\begin{equation}
    \vK(\vt, \vt) - \vK^{01}(\vt, \vs)\vK^{11}(\vs, \vs)^{-1}\vK^{10}(\vs, \vt).
\end{equation}

In addition, we can also make predictions of function values based on a combined set of function values $\vf(\vt)$ and derivative values $\vf^\prime(\vs)$ and this is the so-called \textit{derivative-enhanced} GP \cite{eriksson2018scaling, padidar2021scaling}. Specifically, we have the following joint distribution of $\vf(\vt)$, $\vf^\prime(\vs)$, and function value $\vf(\vu)$ at $\vu$:

\begin{equation}
    \begin{bmatrix}
        \vf(\vt)\\
        \vf^\prime(\vs)\\
        \vf(\vu)
    \end{bmatrix} \sim \mathcal{N}\left(
        \begin{bmatrix}
            \vmu(\vt)\\
            \vmu^\prime(\vs)\\
            \vmu(\vu)
        \end{bmatrix}
        ,
        \begin{bmatrix}
            \vK(\vt, \vt) & \vK^{01}(\vt, \vs) & \vK(\vt, \vu)\\
            \vK^{10}(\vs, \vt) & \vK^{11}(\vs, \vs) & \vK^{10}(\vs, \vu)\\
            \vK(\vu, \vt) & \vK^{01}(\vu, \vs) & \vK(\vu, \vu)
        \end{bmatrix}
    \right).
\end{equation}
We can then make estimation of $\vf(\vu)$ using the following predictive mean:

\begin{equation}
\vmu(\vu) +
\begin{bmatrix}
    \vK(\vu,\vt) & \vK^{01}(\vu,\vs)
\end{bmatrix}
\begin{bmatrix}
    \vK(\vt,\vt) & \vK^{01}(\vt,\vs) \\
    \vK^{10}(\vs,\vt) & \vK^{11}(\vs,\vs)
\end{bmatrix}^{-1}
\begin{bmatrix}
    \vf(\vt) - \vmu(\vt) \\
    \vf^\prime(\vs) - \vmu^\prime(\vs)
\end{bmatrix},\label{eq:derivative-enhanced-gp-predictive-mean}
\end{equation}
with covariance:
\begin{equation}
\vK(\vu,\vu) - 
\begin{bmatrix}
    \vK(\vu,\vt) & \vK^{01}(\vu,\vs)
\end{bmatrix}
\begin{bmatrix}
    \vK(\vt,\vt) & \vK^{01}(\vt,\vs) \\
    \vK^{10}(\vs,\vt) & \vK^{11}(\vs,\vs)
    \end{bmatrix}^{-1}
\begin{bmatrix}
    \vK(\vt,\vu) \\
    \vK^{10}(\vs,\vu)
\end{bmatrix}.\label{eq:derivative-enhanced-gp-predictive-covariance}
\end{equation}

\subsection{An illustrative example}
\begin{figure}[tb]
    \centering
    \begin{subfigure}{0.49\textwidth}
        \includegraphics[width=\textwidth]{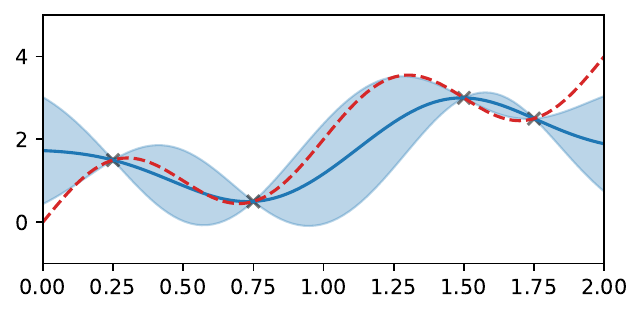}
        \caption{Derivative-free GP}
    \end{subfigure}
    \hfill
    \begin{subfigure}{0.49\textwidth}
        \includegraphics[width=\textwidth]{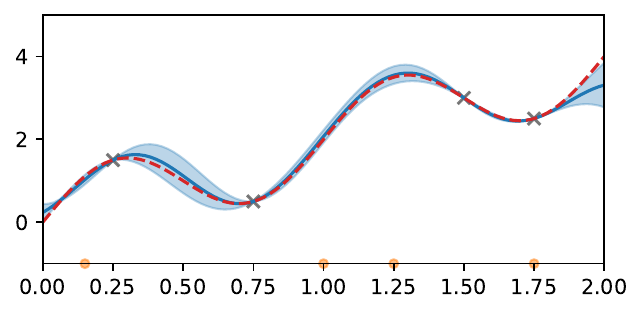}
        \caption{Derivative-enhanced GP}
    \end{subfigure}
    \caption{Derivative-free and derivative-enhanced GPs that approximate the function described in \eqref{eq:example_function}. The black cross markers are the data points. The red dashed line is the ground truth. The blue line is the mean of the GP model with the shaded area showing the 95\% credible interval. The yellow circle markers are the locations of derivative observations.}
    \label{fig:gp-demo}
\end{figure}
To illustrate the above concept on derivative-free and derivative-enhanced GPs, we build GP models to approximate the following 1D function:
\begin{equation}\label{eq:example_function}
    f(t) = \sin(2\pi t) + 2t, \quad t\in[0, 2].
\end{equation}
Assuming a zero mean function and a squared exponential covariance function, we first build a derivative-free GP model based on the function values at points $\vt=$ $\{$0.25, 0.75, 1.5, 1.75$\}$ and the hyperparameters of the covariance function are chosen to maximize the log marginal likelihood \eqref{eq:log-marginal-likelihood}. If we further have derivative observations at points $\vs=$ $\{$0.15, 1.0, 1.25, 1.5$\}$, we can build a derivative-enhanced GP model based on both the function and the derivative observations. The two built GP models are plotted in Fig.~\ref{fig:gp-demo}. We see that the derivative-enhanced GP can better capture the trend of the function than the derivative-free GP, while producing a narrower prediction uncertainty.

%% file: gp_under_constraints.tex
\section{Gaussian Processes under Monotonicity Constraints}
\label{section:gp-under-constraints}

Although GPs provide a flexible framework for estimating complicated functions and quantifying uncertainty, incorporating constraints such as monotonicity presents significant challenges, as it requires constraining an infinite-dimensional object to satisfy inequalities at every point within a continuous space \cite[Section~3]{SwilerSurveyConstrained2020}. To elaborate, we are concerned with the posterior distribution 
\begin{equation}
    \pi(\vf(\vu)|\vf(\vt)),\label{eq:interested-posterior}
\end{equation}
where we wish to enforce the global constraint
\begin{equation}
    \vf^\prime(\vu)\geq 0 \quad \text{for all } \vu \subset \mathcal{T}.\label{eq:global-monotonicity}
\end{equation}

\subsection{Virtual point-based method}
\label{subsection:virtual-point}
The virtual point-based method, independently proposed in \cite{RiihimakiGaussianProcesses2010} and \cite{DaVeigaGaussianProcess2012}, relaxes the global constraint \eqref{eq:global-monotonicity} by substituting it with a finite set of constraints imposed at \textit{virtual} points. By doing so, we can approximate the global behavior of the constrained GP in a more computationally tractable manner, mitigating the numerical issues inherent in the original infinite-dimensional formulation. Concretely, a finite set of virtual points $\vs \subset \mathcal{T}$ is selected and the monotonicity constraints are imposed at these points, i.e., we require only
\begin{equation}
    \vf^\prime(\vs)\geq 0 \quad \text{for } \vs.\label{eq:local-monotonicity}
\end{equation}
The constraint \eqref{eq:local-monotonicity} is a relaxation of \eqref{eq:global-monotonicity}, as it enforces monotonicity only at the finite set of virtual points $\vs$, rather than globally across the entire domain $\mathcal{T}$. The intuition behind this approach is that, with a sufficiently large and well-distributed set of virtual points, the GP is likely to exhibit approximate monotonicity across $\mathcal{T}$, leveraging the smoothness properties of the GP's covariance function.

To incorporate \eqref{eq:local-monotonicity}, the target distribution \eqref{eq:interested-posterior} can be expressed as the marginal of a joint distribution of $\pi(\vf(\vu),\vf^\prime(\vs)|\vf(\vt))$\footnote{From this moment on, we denote data points by $\vt$, virtual points by $\vs$ and prediction points by $\vu$. }, i.e.,
\begin{equation}
        \pi(\vf(\vu)|\vf(\vt)) = \int\pi(\vf(\vu),\vf^\prime(\vs)|\vf(\vt)) \text{d}\vf^\prime(\vs).
\end{equation}
This joint distribution factorizes as:
\begin{equation}
    \pi(\vf(\vu),\vf^\prime(\vs)|\vf(\vt)) = \pi(\vf(\vu)|\vf^\prime(\vs), \vf(\vt))\pi(\vf^\prime(\vs)|\vf(\vt)).
\end{equation}
Using this factorization, samples from \eqref{eq:interested-posterior} under constraint \eqref{eq:local-monotonicity} can be generated via the following three steps:
\begin{enumerate}
    \item sample $\vf^\prime(\vs)$ based on the given $\vf(\vt)$, and ensure that $\vf^\prime(\vs)\geq 0$;
    \item sample $\vf(\vu)$ based on $\vf(\vt)$ and the $\vf^\prime(\vs)$ drawn in the first step;
    \item marginalize out $\vf^\prime(\vs)$ to obtain the posterior distribution $\pi(\vf(\vu)|\vf(\vt))$.
\end{enumerate}

The primary challenge lies in the first step: drawing samples of $\pi(\vf^\prime(\vs)|\vf(\vt))$ that satisfy the constraint. In the following subsections, we will review two existing methods in the literature that address this challenge, i.e., the \textit{truncated prior} and the \textit{non-Gaussian likelihood} methods, and subsequently propose a new method that overcomes the limitations of the existing methods.

We note that the second and third steps are straightforward. Particularly, step 2 can be performed efficiently with derivative-enhanced GP, where analytical forms of the mean and covariance exist for a given pair of samples of $\vf^\prime(\vs)$ and $\vf(\vt)$, as seen in \eqref{eq:derivative-enhanced-gp-predictive-mean} and \eqref{eq:derivative-enhanced-gp-predictive-covariance}.

\subsection{Truncated prior}
\label{subsection:truncated-prior}
By applying Bayes' theorem, we can write the posterior in the following product form
\begin{equation}
    \pi(\vf^\prime(\vs)|\vf(\vt)) \propto \pi(\vf(\vt)|\vf^\prime(\vs))\pi(\vf^\prime(\vs)),\label{eq:gp-bayes}
\end{equation}
where $\pi(\vf(\vt)|\vf^\prime(\vs))$ is the likelihood and $\pi(\vf^\prime(\vs))$ is some suitable prior distribution of $\vf^\prime(\vs)$. Since our interest is to enforce constraints on the samples of $\vf^\prime(\vs)$, a natural strategy is to choose a prior that ensures the constraints are intrinsically satisfied, as proposed in \cite{DaVeigaGaussianProcess2012}. Specifically, we could choose a prior distribution for $\vf^\prime(\vs)$ that restricts support to the non-negative orthant. Then, the prior is defined as
\begin{equation}
    \vf^\prime(\vs)\sim\mathcal{N}(\mathbf{0}, \vK^{11}(\vs, \vs))\mathbb{I}(\vf^\prime(\vs)\geq 0),\label{eq:truncated-prior}
\end{equation}
where $\vK^{11}(\vs, \vs)$ is the covariance matrix of the GP derivatives evaluated at $\vs$ and $\mathbb{I}(\cdot)$ is the indicator function.

As discussed in subsection \ref{subsec:enhanced}, the joint distribution of $\vf(\vt)$ and $\vf^\prime(\vs)$ follows a multivariate Gaussian distribution, expressed as

\begin{equation}
    \begin{bmatrix}
        \vf(\vt)\\
        \vf^\prime(\vs)
    \end{bmatrix} \sim \mathcal{N}\left(
        \begin{bmatrix}
            \vmu(\vt)\\
            \vmu^\prime(\vs)
        \end{bmatrix}
        ,
        \begin{bmatrix}
            \vK(\vt, \vt) & \vK^{01}(\vt, \vs)\\
            \vK^{10}(\vs, \vt) & \vK^{11}(\vs, \vs)
        \end{bmatrix}
    \right).
\end{equation}
Given a sample of $\vf^\prime(\vs)$, we could make use of this conditional relationship and write down the predictive distribution of $\pi(\vf(\vt)|\vf^\prime(\vs))$ as
\begin{equation}
    \vf(\vt)|\vf^\prime(\vs) \sim \mathcal{N}(\vmu^*(\vt), \vSigma^*(\vt)), \label{eq:gp-likelihood}
\end{equation}
with
\begin{equation}\label{eq:affine_mapping}
    \vmu^*(\vt)=\vmu(\vt)+\vK^{01}(\vt, \vs)\vK^{11}(\vs, \vs)^{-1}(\vf^\prime(\vs)-\vmu^\prime(\vs)),
\end{equation}
and
$$\vSigma^*(\vt)=\vK^{00}(\vt, \vt)-\vK^{01}(\vt, \vs)\vK^{11}(\vs, \vs)^{-1}\vK^{10}(\vs, \vt).$$
We note that the predictive mean $\vmu^*(\vt)$ can be interpreted as an affine mapping from $\vf^\prime(\vs)$ to $\vf(\vt)$, and the uncertainty of this mapping is captured by the covariance matrix $\vSigma^*(\vt)$. For the remainder of this paper, we will assume that $\vmu(\vt) = \vmu^\prime(\vs) = \vzero$ for brevity such that \eqref{eq:affine_mapping} simplifies to the linear mapping:
\begin{equation}
    \vmu^*(\vt)= \vA\vf^\prime(\vs), \text{ with } \vA := \vK^{01}(\vt, \vs)\vK^{11}(\vs, \vs)^{-1}.\label{eq:linear_mapping}
\end{equation}

With the above discussed prior and likelihood, the posterior can be expressed as
\begin{equation}
    \begin{aligned}
        \pi(\vf^\prime(\vs)|\vf(\vt)) &\propto {\pi(\vf(\vt)|\vf^\prime(\vs))}\pi(\vf^\prime(\vs))\\
        &= \exp\left(-\frac{1}{2}\|\vA\vf^\prime(\vs)-\vf(\vt)\|^2_{\vSigma^{*}(\vt)^{-1}}-\frac{1}{2}\|\vf^\prime(\vs)\|^2_{\vK^{11}(\vs, \vs)^{-1}}\right)\mathbb{I}(\vf^\prime(\vs)\geq 0),
    \end{aligned}\label{eq:truncated-posterior}
\end{equation}
where the notation $\|\vz\|^2_{\vM^{-1}}$ denotes the squared Mahalanobis norm, i.e., $\vz^\top \vM^{-1} \vz$.

Since the prior \eqref{eq:truncated-prior} is a truncated Gaussian and the likelihood \eqref{eq:gp-likelihood} is Gaussian, it means that the posterior \eqref{eq:truncated-posterior} in this case will be a truncated Gaussian distribution, owing to conjugacy. Note that, for a truncated Gaussian distribution, although the mean and covariance of its associated Gaussian distribution are available, the corresponding moments of the truncated distribution itself do not admit closed-form expressions. Consequently, to estimate statistical properties of the posterior distribution, it is necessary to resort to sampling methods to draw samples from the distribution.
\subsubsection{Sampling methods}
To sample from the truncated Gaussian distribution, several methods are available. The most straightforward approach is rejection sampling. Specifically, we can draw samples from its associated Gaussian distribution and reject the samples that violate the constraints $\vf^\prime(\vs)\geq 0$. This method is simple and easy to implement, but it is computationally expensive as the rejection rate can be significantly high in high-dimensional problems. An alternative method is to use a Gibbs sampler by exploiting the conditional relationship between elements in $\vf^\prime(\vs)$. This method is more efficient than the rejection sampling method, and has been used extensively on constraining GPs with virtual points \cite[e.g.][]{WangEstimatingShape2016, DaVeigaGaussianProcess2020}. Specifically, the Gibbs sampler draws samples of each element in $\vf^\prime(\vs)$ one by one, conditioned on the other elements. This method is efficient in low-dimensional settings, but may still struggle when variables are strongly correlated or when the number of virtual points is large.

We note that the truncated Gaussian distribution can be transformed into an unconstrained distribution by using a change of variables. Specifically, we can transform $\vf^\prime(\vs)$ with the following component-wise mapping:
\begin{equation}
    [\vf^\prime(\vs)]_i = \exp(\vx_i),\label{eq:transformation-change-of-variables}
\end{equation}
where $\vx$ follows an unconstrained distribution. The posterior density of $\vx$ can be obtained through a change of variables formula:
\begin{equation}
    \pi(\vx|\vf(\vt)) = \pi(\vf^\prime(\vs)|\vf(\vt))\left|\det\left(\frac{\partial \vf^\prime(\vs)}{\partial \vx}\right)\right|,
\end{equation}
where $\det(\partial \vf^\prime(\vs)/\partial \vx)$ is the determinant of the Jacobian matrix of the transformation, which for \eqref{eq:transformation-change-of-variables} equals $\prod_{i = 1}^{n}\exp(\vx_i)$.

By doing so, it allows us to use gradient-based samplers to efficiently draw samples of $\vx$ in an unconstrained space. In this work, we will use NUTS to draw samples from the transformed distribution. NUTS is a Hamiltonian Monte Carlo (HMC) sampler that is efficient in sampling from high-dimensional distributions. Nevertheless, we note that NUTS is a MCMC-based sampling method and the samples are correlated and it also requires a burn-in period to tune the parameters of the sampler. Applying \eqref{eq:transformation-change-of-variables} to the drawn samples of $\vx$ will give us back the samples of $\vf^\prime(\vs)$ from our desired posterior.

\subsection{Non-Gaussian likelihood}
\label{subsection:non-gaussian-likelihood}
The method reviewed in subsection~\ref{subsection:truncated-prior} will lead to a strictly increasing function on the virtual points. This is due to the fact that the constructed posterior $\pi(\vf^\prime(\vs)|\vf(\vt))$ is a truncated Gaussian distribution and the possibility of $\vf^\prime(\vs)=0$ is zero. However, we may want to allow the function to have zero gradients, as it is natural, and even desirable in some cases, to have a flat region in a monotonic function.

To address this issue, \cite{WangEstimatingShape2016} proposed a method that forms the posterior distribution in a different way, which we will refer to as the \textit{non-Gaussian likelihood} method. In \cite{WangEstimatingShape2016}, a ReLU transformation is applied to $\vf^\prime(\vs)$ to get
\begin{equation}
    \text{ReLU}(\vf^\prime(\vs))_i = \max(\vf^\prime(\vs), 0),
\end{equation}
which is an element-wise transformation that sets the negative values to zero. $\text{ReLU}(\vf^\prime(\vs))$, in place of $\vf^\prime(\vs)$, is then used to predict $\vf(\vt)_i$ and then evaluate the likelihood, so
\begin{equation}
    \begin{bmatrix}
        \vf(\vt)\\
        \text{ReLU}(\vf^\prime(\vs))
    \end{bmatrix} \sim \mathcal{N}\left(
        \begin{bmatrix}
            \vmu(\vt)\\
            \vmu^\prime(\vs)
        \end{bmatrix}
        ,
        \begin{bmatrix}
            \vK(\vt, \vt) & \vK^{01}(\vt, \vs)\\
            \vK^{10}(\vs, \vt) & \vK^{11}(\vs, \vs)
        \end{bmatrix}
    \right),
\end{equation}
By doing so, the method allows the possibility of $\text{ReLU}(\vf^\prime(\vs))$ being exactly zero, i.e., when $\vf^\prime(\vs)\leq 0$, and the posterior can be written as

\begin{equation}
    \begin{aligned}
        \pi(\vf^\prime(\vs)|\vf(\vt)) &\propto {\pi(\vf(\vt)|\vf^\prime(\vs))}\pi(\vf^\prime(\vs))\\
        &= \exp\left(-\frac{1}{2}\|\vA(\boldsymbol{\text{ReLU}}(\vf^\prime(\vs)))-\vf(\vt)\|^2_{\vSigma^{*}(\vt)^{-1}}-\frac{1}{2}\|\vf^\prime(\vs)\|^2_{\vK^{11}(\vs, \vs)^{-1}}\right),
    \end{aligned}\label{eq:non-gaussian-posterior}
\end{equation}
where $\vA$ is the same linear operator as defined in \eqref{eq:linear_mapping}, but now it is used to map $\text{ReLU}(\vf^\prime(\vs))$ to $\vf(\vt)$. It is noted that while the operator $\vA$ provides a linear mapping from $\text{ReLU}(\vf^\prime(\vs))$ to $\vf(\vt)$, the application of the $\text{ReLU}$ function introduces nonlinearity into the model. As a result, the likelihood function becomes non-Gaussian, and consequently, the posterior distribution no longer retains a Gaussian form. This nonlinearity complicates both the analytical characterization and the sampling of the posterior.

\subsubsection{Sampling methods}
In \cite{WangEstimatingShape2016}, the authors proposed to use a component-wise Gibbs sampler to draw samples from the posterior \eqref{eq:non-gaussian-posterior}. We note that the Gibbs sampler in \cite{WangEstimatingShape2016} was specifically designed for this non-Gaussian likelihood model and does not readily generalize to more general transformations than ReLU. Alternatively, since the posterior density gradient is available almost everywhere, gradient-based samplers can be employed. In this work, we again propose to use NUTS, which does not require problem-specific modifications and is more straightforward to implement than the Gibbs sampler.

\subsection{Regularized Linear Randomize-then-Optimize-based method}
Despite the effectiveness of the two methods reviewed in subsection~\ref{subsection:truncated-prior} and subsection~\ref{subsection:non-gaussian-likelihood}, each has notable limitations. The truncated prior approach enforces strictly positive gradients, thereby excluding the possibility of flat regions in the function. The non-Gaussian likelihood method permits zero gradients, but still relies on MCMC-based sampling, which can be computationally intensive, especially as the number of virtual points increases. Additionally, MCMC samples are typically correlated and require careful tuning, such as burn-in and thinning, to ensure sample quality.

With the above issues in view, we naturally raise a question: is there an alternative method that might overcome the difficulties of the existing methods? Here are some properties of the new method that we desire:
\begin{itemize}
    \item \textbf{Ability to have zero gradients}. We would like to have a method that allows the function to have zero gradients.
    \item \textbf{Computational efficiency}. The method should be computationally efficient for even large scale problems. In constraining the GP, we might need to put a large number of virtual points, and the method should be able to handle this efficiently.
    \item \textbf{Simplicity and robustness}. The method should be easy to implement and preferably make use of pre-existing, black-box sampling methods.
\end{itemize}

\subsubsection{Regularized Linear Randomize-then-Optimize method}

In this work, we propose to constrain GPs with the RLRTO method that enjoys the desirable properties outlined above. The Randomize-then-Optimize (RTO) method was proposed as an efficient method to draw samples in high-dimensional inverse problems \cite{bardsley2014randomize}. When the likelihood is linear Gaussian and the prior is Gaussian, RTO demonstrates particular efficiency with each sample drawn being independent \cite{bardsley2012RTO}. Applying this method the unconstrained, derivative-enhanced GP setting, results in sampling from the derivative GP by repeatedly solving the following \textit{unconstrained} linear least squares problem:
\begin{equation}\label{eq:linear_RTO}
    \argmin_{\vf^\prime(\vs)}\frac{1}{2}\|\vA\vf^\prime(\vs)-\hat{\vb}\|^2_{{\vSigma^{*}(\vt)}^{-1}} + \frac{1}{2}\|\vf^\prime(\vs)-\hat{\vc}\|^2_{\vK^{11}(\vs, \vs)^{-1}},
\end{equation}
where $\hat{\vb}$ is a sample from $\mathcal{N}(\vf(\vt), {\vSigma^{*}(\vt)})$ and $\hat{\vc}$ is a sample from $\mathcal{N}(\mathbf{0}, \vK^{11}(\vs, \vs))$.

As we usually have well-established algorithms to solve optimization problems like \eqref{eq:linear_RTO}, e.g., Krylov subspace methods, even in high-dimensional settings, each perturbed optimization can be solved relatively efficiently. Particularly, the numerical implementation is straightforward, as seen in Algorithm~\ref{alg:rto-gaussian}, and can readily make use of existing solver libraries. In addition, the samples generated by the RTO method are independent, which is desirable in practice, and no burn-in period is required. These features make the RTO method rather promising when compared with MCMC-based methods.

\begin{algorithm}[tb]
    \begin{algorithmic}
    \STATE{$i=0$}
    \WHILE{$i < N$}
    \STATE{Generate $\hat{\vb}$ from $\mathcal{N}(\vf(\vt), \vSigma^*(\vt))$ and $\hat{\vc}$ from $\mathcal{N}(\mathbf{0}, \vK^{11}(\vs, \vs))$}
    \STATE{Solve 
        $\argmin_{\vf^\prime(\vs)}\frac{1}{2}\|\vA\vf^\prime(\vs)-\hat{\vb}\|^2_{{\vSigma^{*}(\vt)}^{-1}} + \frac{1}{2}\|\vf^\prime(\vs)-\hat{\vc}\|^2_{\vK^{11}(\vs, \vs)^{-1}}$}
    \STATE{Store the solution as a sample}
    \STATE{$i=i+1$}
    \ENDWHILE
    \end{algorithmic}
    \caption{RTO for unconstrained GPs}
    \label{alg:rto-gaussian}
\end{algorithm}

The RTO method has since been extended to RLRTO to accommodate posterior samples under non-negativity constraint in \cite{bardsley2012mcmc} and \cite{bardsley2020mcmc}. In our case of drawing constrained samples of $\vf^\prime(\vs)$, RLRTO works by repeatedly solving the following \textit{constrained} optimization problem:
\begin{equation}
    \argmin_{\vf^\prime(\vs) \geq 0}\frac{1}{2}\|\vA\vf^\prime(\vs)-\hat{\vb}\|^2_{{\vSigma^{*}(\vt)}^{-1}} + \frac{1}{2}\|\vf^\prime(\vs)-\hat{\vc}\|^2_{\vK^{11}(\vs, \vs)^{-1}}.\label{eq:rto-nonnegativity}
\end{equation}
Since the non-negativity constraint forms a convex feasible region and the objective function \eqref{eq:rto-nonnegativity} is strictly convex, the existence and uniqueness of the solution are guaranteed. The implementation remains straightforward, being nearly identical to Algorithm~\ref{alg:rto-gaussian}, but with the addition of the non-negativity constraints. The solution of \eqref{eq:rto-nonnegativity} can be determined using a variety of efficient algorithms, e.g., the active set, interior-point, and projected gradient methods. The solutions from the perturbed optimization problems correspond to samples of a desired posterior distribution, effectively balancing the impact of the measurement and the prior while strictly enforcing the non-negativity constraint.

We note that the samples generated from \eqref{eq:rto-nonnegativity} can be interpreted from at least two different perspectives. From the first perspective, the RLRTO method can be seen as a method that draws samples from the unconstrained posterior, i.e., by solving \eqref{eq:linear_RTO}, and then projects the samples to the feasible set, $\vf^\prime(\vs) \geq 0$ in this case, with respect to a particular norm; see \cite{bardsley2012mcmc} for details. By solving \eqref{eq:rto-nonnegativity}, there is no need to form this norm or perform the projection explicitly. From the second perspective, the RLRTO method can be seen as a method that enforces an \textit{implicit} prior, whose density does not have a closed form, but is implicitly imposed through the process of solving the constrained optimization problem. As a result, the posterior can not be written in a conventional Bayesian sense, i.e., in a form like \eqref{eq:non-gaussian-posterior}, though we can still efficiently draw samples from it.

We would like to emphasize that the solution of \eqref{eq:rto-nonnegativity} may lie on the boundary of the feasible set, and this is a desired property in this work as it permits the possibility of having zero gradients, i.e., $\vf^\prime(\vs)=0$.

\subsection{Comparison of methods}
Key properties of the truncated prior, the non-Gaussian likelihood and our proposed RLRTO methods are summarized in Table~\ref{tab:comparison}. Whilst the RLRTO approach has the unfavorable property of having no explicit density, it has various favorable properties which can both improve the quality of the constrained GP as well as its computational efficiency.

In addition, the sampling process of the RLRTO method is \textit{embarrassingly parallelizable}, as its samples are independent. This is in contrast to the MCMC-based methods, including Gibbs and NUTS mentioned in this work, where the samples are correlated and the sampling process is sequential. Though not explored here, this inherent parallelization is particularly beneficial when dealing with a large number of virtual points, enabling more efficient posterior sampling.

\begin{table}
    \centering
\caption{Comparison of methods for building GP models under monotonicity constraints.}
\begin{tabular}{ccccc}
    \hline
    Methods & Truncated prior & Non-Gaussian likelihood & RLRTO (ours) \\
    \hline 
    Density & Explicit \eqref{eq:truncated-posterior} & Explicit \eqref{eq:non-gaussian-posterior} & Implicit \\
    Sampling & MCMC & MCMC & Optimization \\
    Samples & Correlated & Correlated & Independent \\
    Allows $\vf^\prime(\vs)=0$? & No & Yes & Yes \\
    \hline
    \label{tab:comparison}
\end{tabular}
\end{table}

\subsection{Constraints on higher-order derivatives}
We note that this section so far focuses on constraining the first-order derivatives of the function. However, in some cases, we might be interested in constraining higher-order derivatives of the function. We note that all three methods, i.e., truncated prior, non-Gaussian likelihood, and RLRTO, can be used to apply inequality constraints to higher-order derivatives. For example, we can use the constraint of $\vf^{\prime\prime}(\vs)\geq 0$ to force the approximated function to be convex. In this case, the joint distribution of $\vf(\vt)$ and $\vf^{\prime\prime}(\vs)$ can be written in a similar way as \eqref{eq:gp-likelihood} to obtain the corresponding forward model $\vA$. The sampling procedures will stay the same for each of the three methods. However, when considering convexity constraints in higher dimensions, the component-wise non-negativity constraints generalize to requiring the Hessians to be positive semidefinite. Due to the additional complexity of such a quadratic constraint, both in terms of notation and sampling, we leave this as future work.

\subsection{Locations of virtual points}
The locations of the virtual points $\vs$ are important in constraining the GP. In this work, we will use a set of virtual points that are pre-determined by sampling from the Sobol sequence \cite{sobol1967distribution}, which provides a well-distributed coverage of the domain. Alternatively, the locations could be dynamically added at each iteration based on the current samples of $\vf(\vt)$ and $\vf^\prime(\vs)$. This adaptive strategy is similar to the active learning method, where the locations of the virtual points are adaptively chosen based on the current samples, \cite[e.g.][]{WangEstimatingShape2016}. However, such a method leads to many rounds of sampling and increases computational cost. In this work, we will not consider this adaptive method, but we note that it is a possible future direction.

%% file: experiments.tex
\section{Experiments}
\label{section:experiments}

The purpose of this section is to show the effect of constraining a GP with the methods discussed in Section~\ref{section:gp-under-constraints}, including the truncated prior, the non-Gaussian likelihood and the RLRTO model. We choose to build constrained GPs for known functions, so we can easily evaluate the performance of the models both in terms of quality and computational speed.

\subsection{Implementation details}\label{subsec:implementations}
To make the comparisons in computation time fair, we have made use of trusted, pre-implemented samplers and optimizers, when available. In particular, we have made use of the following implementations of the samplers and optimizers:
\begin{itemize}
    \item \textbf{Gibbs for truncated prior and non-Gaussian likelihood:} Our own implementation of the component-wise Gibbs samplers proposed in \cite{WangEstimatingShape2016}.
    \item \textbf{NUTS for truncated prior and non-Gaussian likelihood:} We use \texttt{CUQIpy} \cite{riis2024cuqipy, alghamdi2024cuqipy} to model posterior \eqref{eq:non-gaussian-posterior} with NUTS implemented in \texttt{Pyro} \cite{bingham2019pyro}.
    \item \textbf{RLRTO:} We use the RLRTO method implemented in \texttt{CUQIpy}, which solves the constrained linear least squares problem using L-BFGS-B from \texttt{SciPy} \cite{2020SciPy-NMeth}.
\end{itemize}

For each method, we draw 51,000 samples of $\vf^\prime(\vs)$, discard the first 1,000 samples as burn-in, and use the remaining $N = 50,000$ samples to obtain samples from $\vf(\vu)$. Withe these samples, we then build constrained GP models based on the steps listed in subsection~\ref{subsection:virtual-point} and also an unconstrained GP model for comparison.

\subsection{Evaluation metrics}\label{subsec:metrics}
For a proper comparison between the unconstrained and various constrained GPs, we use various metrics that summarize either the quality of the approach or the efficiency of the sampling method. In particular, to measure the quality of the GP, we use the following quantities:
\begin{itemize}
    \item \textbf{Mean Square Error (MSE):} Otherwise referred to as the empirical risk and defined by $\frac{1}{N}\sum_{i=1}^N\|\vf_i - \vf_{\text{true}}\|_2^2$. It measures the overall error of the GP model. A lower MSE implies a better fit to the ground truth.
    \item \textbf{Average $95\%$ credible interval width (CI width): } The average of the component-wise $95\%$ credible interval width measures the total amount of uncertainty in the GP. A lower CI width implies a stronger subjective belief, but could also signify overfitting or overconfidence in the model.
\end{itemize}

To measure the computational efficiency of the sampling method, we use the following quantities:
\begin{itemize}
    \item \textbf{Average integrated autocorrelation time $\tau_{\text{aut}}$ (IAT): } The average of the component-wise integrated autocorrelation times describes the number of possibly correlated samples needed to get the equivalent of a single independent sample. Related to the effective sample size (ESS) $N_{\text{ess}}$ by $\tau_{\text{aut}} = N/N_{\text{ess}}$. This quantity should only depend on the algorithm and not the actual implementation and hardware.
    \item \textbf{Effective samples per second (ESS per second): } The number of effective samples per second, i.e., $N_{\text{ess}}/t_{\text{runtime}}$. It is an implementation and hardware specific descriptor of the efficiency.  
\end{itemize}

\subsection{Experiment setup}
The experiments were conducted on six functions, summarized in Table~\ref{table:experiment-expressions}. The first three are 1D functions defined on $[-5, 5]$, while the last three are 2D functions defined on $[-5, 5] \times [-5, 5]$. These functions were chosen to represent diverse characteristics:
\begin{itemize}
    \itemsep0em 
    \item \textbf{1D-1:} Logarithmic function with sparse data with only four observations.
    \item \textbf{1D-2:} Piecewise linear function with a flat region lacking observations.
    \item \textbf{1D-3:} Function with many noisy observations.
    \item \textbf{2D-1:} Sinusoidal function with a zero gradient in the first direction.
    \item \textbf{2D-2:} Logarithmic function with a strictly positive gradient in the first direction.
    \item \textbf{2D-3:} Function with both zero and positive gradients in the first direction.
\end{itemize}

Other important parameters of the experiments are summarized in Table~\ref{table:experiment-expressions}. The number of observations is the number of data points used to train the GP models. The noise level is the standard deviation of the Gaussian noise added to the function values. In each experiment, we build a set of GP models using varying numbers of virtual points $\{$4, 8, 16, 32, 64, 128$\}$. Note that the number of virtual points is only relevant to the constrained GP models.

\begin{table}[tb]
    \centering
    \caption{Function expressions for experiments with ground truths, together with the number of observations and the noise level. For the 1D examples, the single parameter is $t$, whilst for the 2D examples, the two parameters are $t_1$ and $t_2$.}
    \begin{tabular}{l c c c} 
     \hline
     Index & Expression & Observations & Noise level \\
     \hline
     1D-1 & $\log(t+5.1)$ & 4 & $10^{-1}$\\
     1D-2 & $\begin{cases} 
        t+3 &  t< -3 \\
        0 & -3\leq t < 3 \\
        t-3 & t \geq 3
     \end{cases}$ & 64 & $10^{-1}$\\
     1D-3 & $\displaystyle\frac{4}{1+\exp(4-t)}$ & 50 & $3\cdot10^{-1}$\\
     2D-1 & $\sin (t_2)$& 16 & $10^{-3}$\\
     2D-2 & $\log(t_1+ 6)(1-\cos(t_2))$ & 16 & $10^{-3}$\\
     2D-3 & $t_1\cos^2(t_2)$ & 64 & $10^{-3}$\\
     \hline
    \end{tabular}
    \label{table:experiment-expressions}
\end{table}

The noise-free data are generated by evaluating the function expressions in Table~\ref{table:experiment-expressions} at the data points $\vt$. The data are then corrupted by adding Gaussian noise with a certain standard deviation to the function values. The data is then used to train the GP models. The locations of the data points are sampled from Latin hypercubes \cite{mckay2000comparison}, except for the 1D-1 case, which is manually chosen. Meanwhile, the locations of the virtual points $\vs$ are sampled using randomized Sobol' sequences, such that we can consider sequentially adding more virtual points. The locations of the test points $\vu$ are chosen to be equally spaced in the domain of interest.

The hyperparameters of the GP models are optimized by maximizing the negative log-likelihood defined in \eqref{eq:log-marginal-likelihood}. Specifically, we use the stochastic gradient descent (SGD) optimizer through \texttt{GPJax} \cite{pinder2022gpjax} with a learning rate of $0.01$ and a maximum of 20,000 iterations. Once optimized, the same hyperparameters are used to build both the unconstrained and constrained GP models. In other words, in each experiment, e.g., 1D-1, the hyperparameters are optimized only once, and the same hyperparameters are used for all the GP models.

The unconstrained GP model is built only with the data points and its mean and covariance matrix are analytically available. The constrained GP models are built with the data points whilst enforcing monotonicity (only in the first dimension in 2D experiments) at the virtual points.

\subsection{Experiment results}

\begin{figure}[H]
    \centering
    \includegraphics[width=\linewidth]{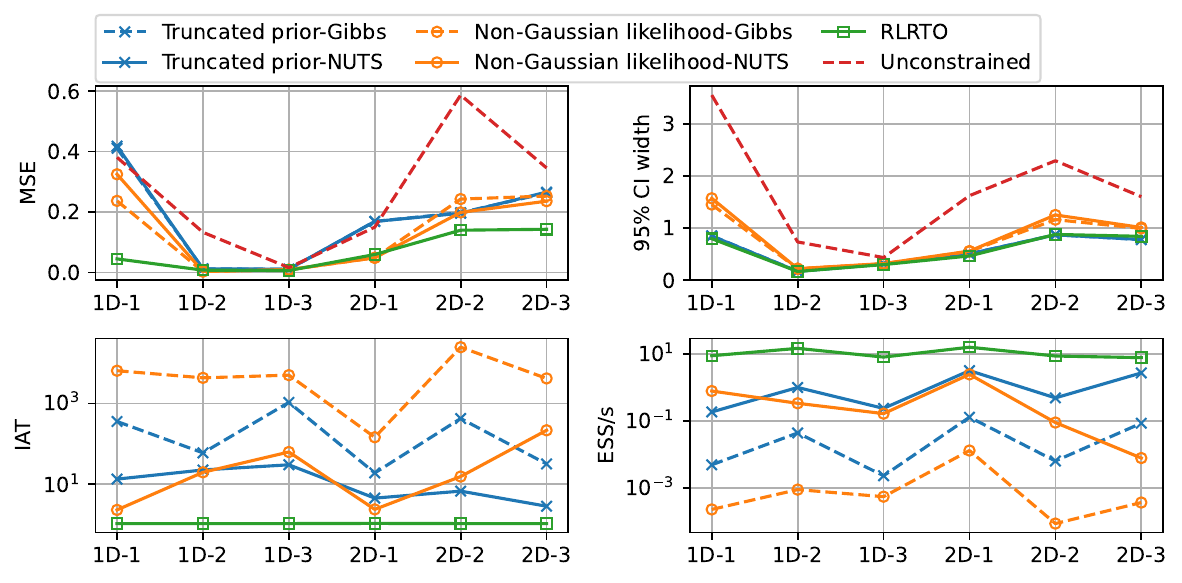}
    \caption{Summary of metrics of GP models built for the 1D and 2D experiments as specified in Table~\ref{table:experiment-expressions} with 128 virtual points.}
    \label{fig:metrics_against_experiments}
\end{figure}

We plot the constrained GP models built with the highest number of virtual points, namely 128, against the unconstrained model in Figs.~\ref{fig:experiment-1d}, \ref{fig:experiment-2d-mean}, and \ref{fig:experiment-2d-ci}. Fig.~\ref{fig:experiment-1d} combines predictive means and 95\% CI width for 1D cases, while Figs.~\ref{fig:experiment-2d-mean} and \ref{fig:experiment-2d-ci} separately show predictive means and 95\% CI width for 2D cases. Fig.~\ref{fig:metrics_against_experiments} summarizes the metrics of these GP models.

Note that all constrained approaches greatly improve the quality of the GP over the unconstrained GP, including: removing any non-monotonicity, more similarity to the ground truth, and a reduction in the size of the 95\% CI width. Furthermore, in all the case where monotonicity is enforced whilst the gradient is (near-)zero, the nonlinear and RLRTO approaches outperform the truncated approach, due to the truncated approach trying to enforce a strictly increasing signal.

Whilst the difference between the unconstrained and constrained GPs are clearly visible, the differences between the different constrained GPs can be more subtle. Hence it is preferred to do a quantitative over a qualitative comparison.

\input{examples_1d}

\input{examples_2d_mean}
\input{examples_2d_CI}

\begin{figure}[tb]
    \centering
    \includegraphics[width=0.9\linewidth]{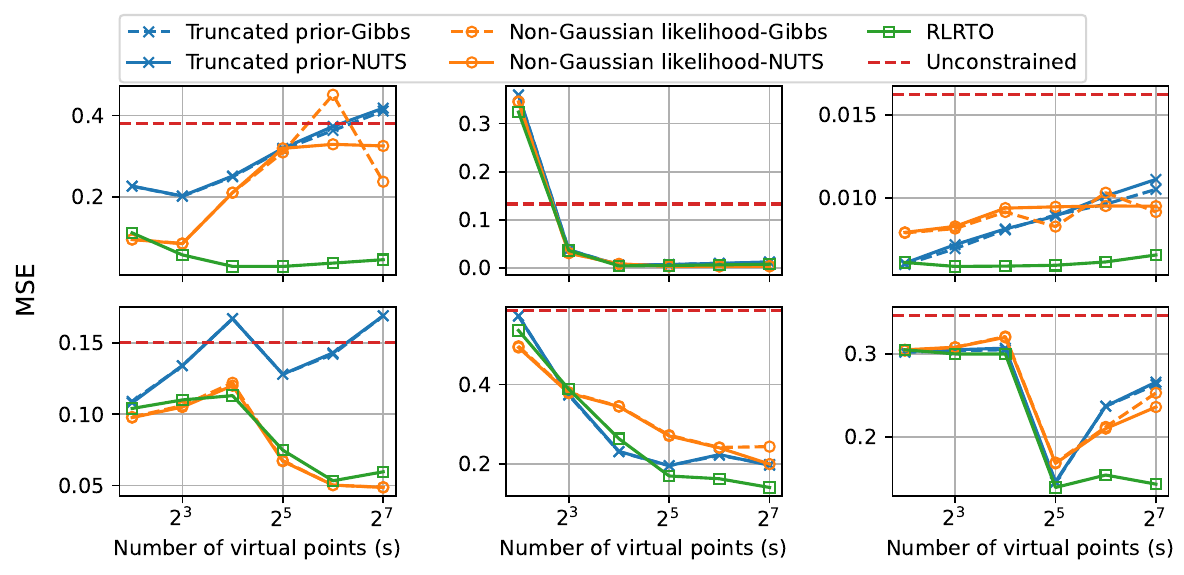}
    \caption{MSE of the GPs as the number of virtual points increases. The first row corresponds the 1D experiments and second row the 2D ones.}
    \label{fig:MSE_over_virtual_points}
\end{figure}

\paragraph{Quality metrics}
The MSEs of the (un)constrained GPS are shown in Fig.~\ref{fig:MSE_over_virtual_points} with a variable number of virtual points, and in the top left panel of Fig.~\ref{fig:metrics_against_experiments} for 128 virtual points. We can see that all constrained GPs improve upon the unconstrained GP, with RLRTO being generally the best. When we increase the number of virtual points, we see that a relatively small number of virtual points already improves the MSE a lot, whilst increasing the number of virtual points need not keep improving the MSE. Note that the MSE is not expected to go to zero as the number of virtual points increase, as we are not actually providing more observations, but only restricting the search space of the process.

\begin{figure}[tb]
    \centering
    \includegraphics[width=0.9\linewidth]{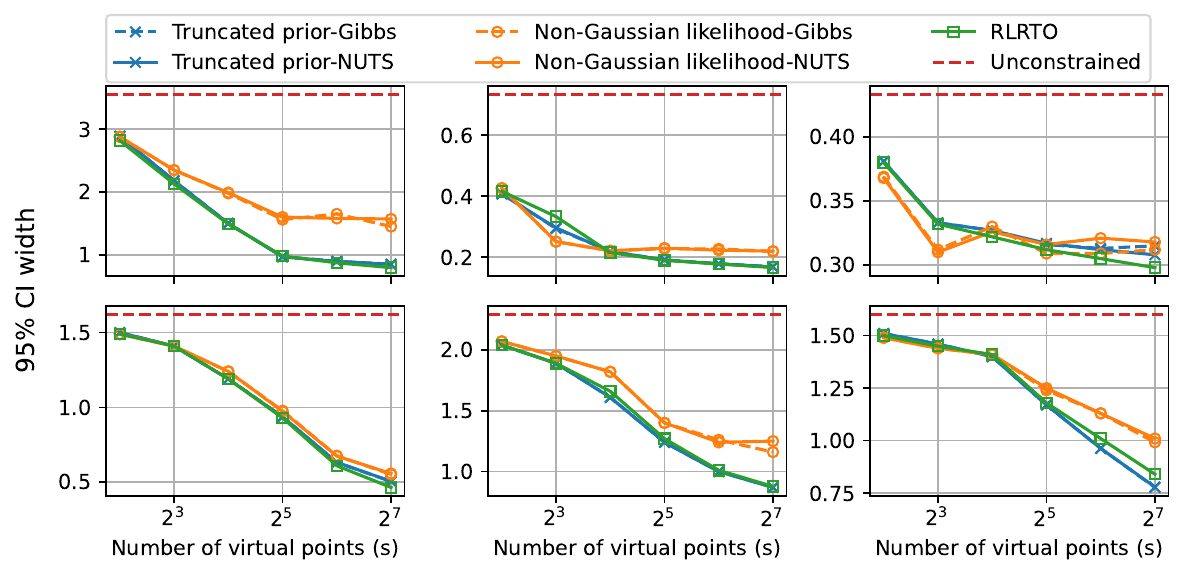}
    \caption{Average CI width of the GPs as the number of virtual points increases. The first row corresponds to the 1D experiments and the second row to the 2D ones.}
    \label{fig:CI_over_virtual_points}
\end{figure}

The average 95\% CI widths of the (un)constrained GPS are shown in Fig.~\ref{fig:CI_over_virtual_points} with a variable number of virtual points, and in top right panel of Fig.~\ref{fig:metrics_against_experiments} for 128 virtual points. When incorporating constraints, the search space of the process becomes smaller, and thus the total uncertainty is expected to be smaller. Indeed, it is shown that the average 95\% CI width is smaller in all the experiments. The average CI width is often very similar for all the constrained approaches.

\paragraph{Efficiency metrics}

\begin{figure}[tb]
    \centering
    \includegraphics[width=0.9\linewidth]{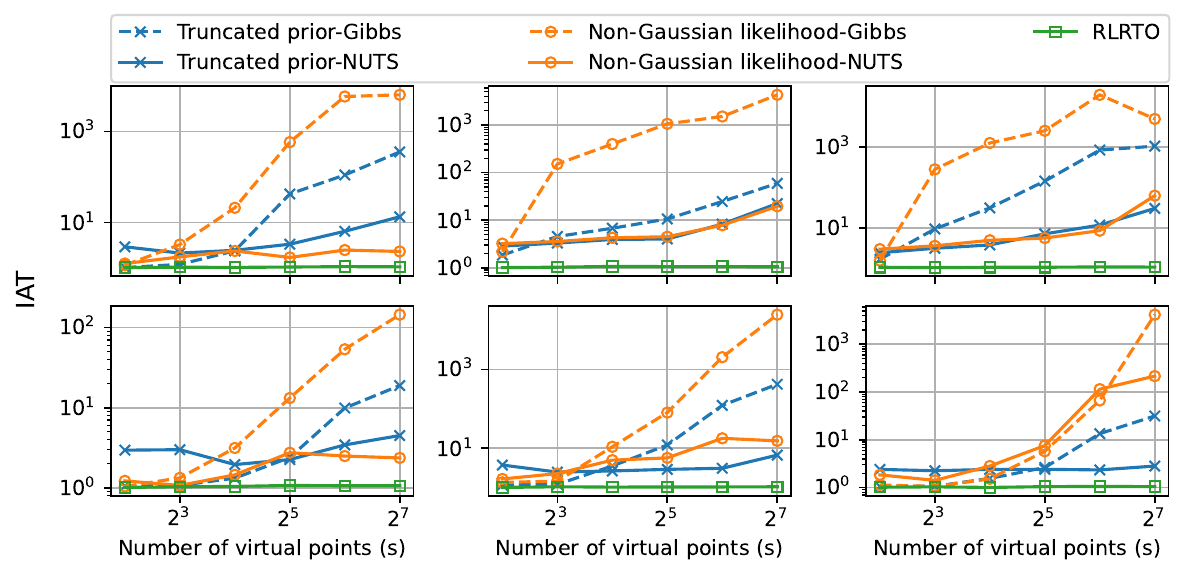}
    \caption{Average IAT of the constrained GPs as the number of virtual points increases. The first row corresponds to the 1D experiments and the second row to the 2D ones.}
    \label{fig:IAT_over_virtual_points}
\end{figure}

The average IATs of the constrained GPS are shown in Fig.~\ref{fig:IAT_over_virtual_points} with a variable number of virtual points, and in the bottom left panel of Fig.~\ref{fig:metrics_against_experiments} for 128 virtual points. Both figures show that the IAT of RLRTO is almost one with some very small perturbations due to correlation introduced by inaccurately solving the constrained linear least squares problem and using the previous sample as warm-start for the next sample. Both figures also show that for a sufficiently big number of virtual points, the IAT for NUTS is lower than for Gibbs. NUTS also seems to scale better than Gibbs in all cases.

\begin{figure}[tb]
    \centering
    \includegraphics[width=0.9\linewidth]{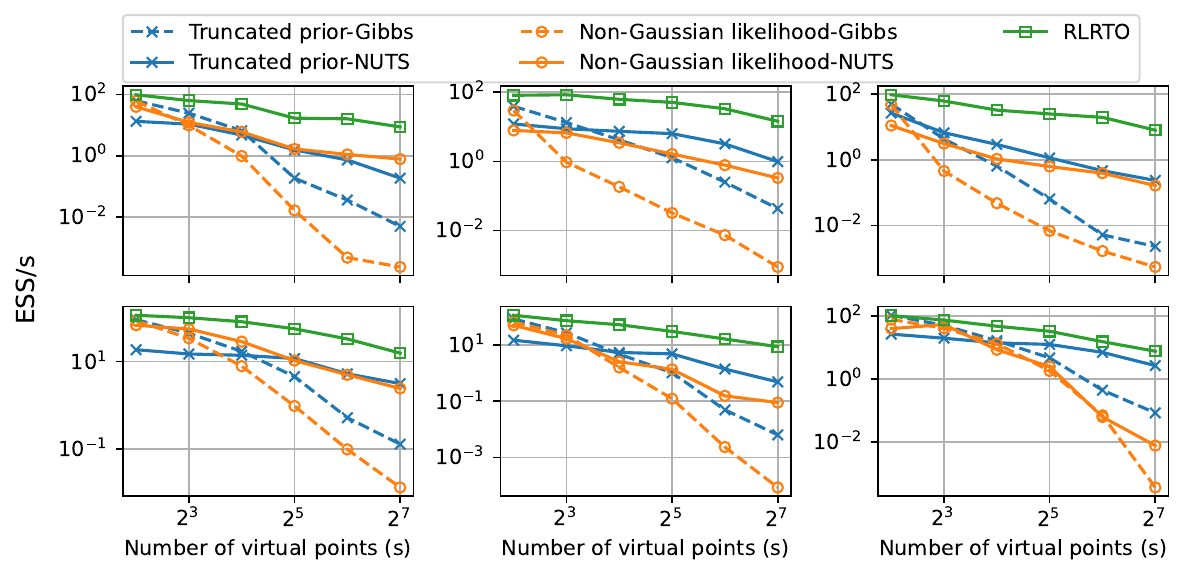}
    \caption{ESS per second of the constrained GPs as the number of virtual points increases. The first row corresponds to the 1D experiments and the second row to the 2D ones.}
    \label{fig:ESS_over_virtual_points}
\end{figure}

The values of ESS per second of the constrained GPS are shown in Fig.~\ref{fig:ESS_over_virtual_points} with a variable number of virtual points, and in the bottom right panel of Fig.~\ref{fig:metrics_against_experiments} for $128$ virtual points. Both Figures show that RLRTO outperforms NUTS, whilst NUTS outperforms Gibbs for a sufficiently large number of virtual points, similar to the behaviour for IAT in Fig.~\ref{fig:metrics_against_experiments}.

It is important to acknowledge that the implementations of RLRTO, NUTS and Gibbs differ in their level of optimization. In particular, the Gibbs implementation could be optimized with further tuning, and more specialized or efficient solvers than L-BFGS-B could be used for solving the constrained linear least squares problem in RLRTO. Nonetheless, based on the scaling behavior of the IAT seen in Fig.~\ref{fig:MSE_over_virtual_points}, it is still expected that RLRTO outperforms NUTS, which in turn outperforms Gibbs, as the number of virtual points increases.

%% file: examples_1d.tex
\begin{figure}[H]
\centering
\begin{subfigure}[t]{\dimexpr0.135\textwidth+10pt\relax}
    \makebox[10pt]{\raisebox{20pt}{\rotatebox[origin=c]{90}{\footnotesize{1D-1}}}}%
    \includegraphics[width=\dimexpr\linewidth-10pt\relax, clip, trim={0.1cm 0.1cm 0.1cm 0.1cm}]
    {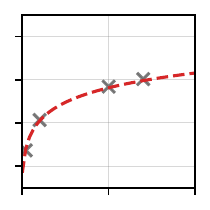}
    \makebox[10pt]{\raisebox{20pt}{\rotatebox[origin=c]{90}{\footnotesize{1D-2}}}}%
    \includegraphics[width=\dimexpr\linewidth-10pt\relax, clip, trim={0.1cm 0.1cm 0.1cm 0.1cm}]
    {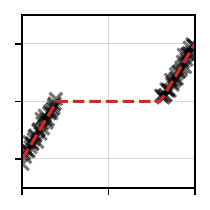}
    \makebox[10pt]{\raisebox{20pt}{\rotatebox[origin=c]{90}{\footnotesize{1D-3}}}}%
    \includegraphics[width=\dimexpr\linewidth-10pt\relax, clip, trim={0.1cm 0.1cm 0.1cm 0.1cm}]
    {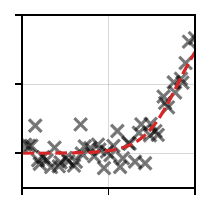}
    \caption*{Ground truth}
\end{subfigure}
\begin{subfigure}[t]{0.135\textwidth}
    \includegraphics[width=\textwidth, clip, trim={0.1cm 0.1cm 0.1cm 0.1cm}]  
    {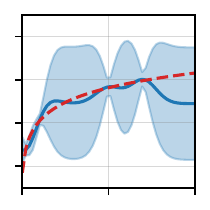}
    \includegraphics[width=\textwidth, clip, trim={0.1cm 0.1cm 0.1cm 0.1cm}]
    {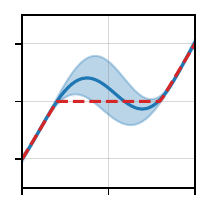}
    \includegraphics[width=\textwidth, clip, trim={0.1cm 0.1cm 0.1cm 0.1cm}]
    {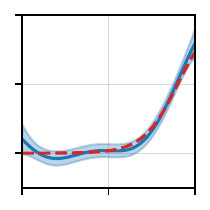}
    \caption*{Unconstrained}
\end{subfigure}
\begin{subfigure}[t]{0.135\textwidth}
    \includegraphics[width=\textwidth, clip, trim={0.1cm 0.1cm 0.1cm 0.1cm}]  
    {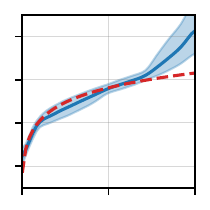}
    \includegraphics[width=\textwidth, clip, trim={0.1cm 0.1cm 0.1cm 0.1cm}]
    {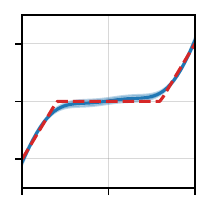}
    \includegraphics[width=\textwidth, clip, trim={0.1cm 0.1cm 0.1cm 0.1cm}]
    {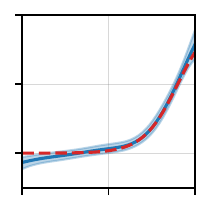}
    \caption*{Truncated prior-Gibbs}
\end{subfigure}
\begin{subfigure}[t]{0.135\textwidth}
    \includegraphics[width=\textwidth, clip, trim={0.1cm 0.1cm 0.1cm 0.1cm}]  
    {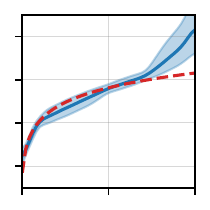}
    \includegraphics[width=\textwidth, clip, trim={0.1cm 0.1cm 0.1cm 0.1cm}]
    {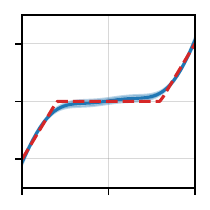}
    \includegraphics[width=\textwidth, clip, trim={0.1cm 0.1cm 0.1cm 0.1cm}]
    {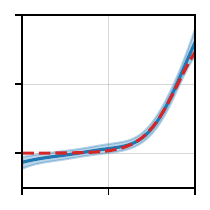}
    \caption*{Truncated prior-NUTS}
\end{subfigure}
\begin{subfigure}[t]{0.135\textwidth}
    \includegraphics[width=\textwidth, clip, trim={0.1cm 0.1cm 0.1cm 0.1cm}]  
    {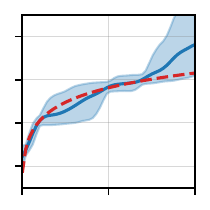}
    \includegraphics[width=\textwidth, clip, trim={0.1cm 0.1cm 0.1cm 0.1cm}]
    {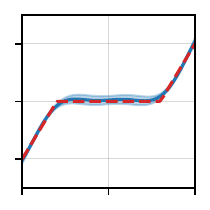}
    \includegraphics[width=\textwidth, clip, trim={0.1cm 0.1cm 0.1cm 0.1cm}]
    {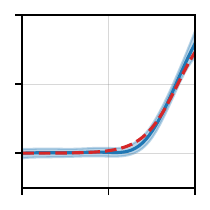}
    \caption*{Non-Gaussian likelihood-Gibbs}
\end{subfigure}
\begin{subfigure}[t]{0.135\textwidth}
    \includegraphics[width=\textwidth, clip, trim={0.1cm 0.1cm 0.1cm 0.1cm}]  
    {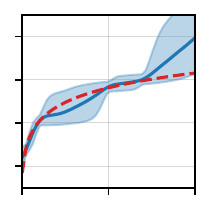}
    \includegraphics[width=\textwidth, clip, trim={0.1cm 0.1cm 0.1cm 0.1cm}]
    {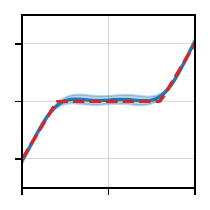}
    \includegraphics[width=\textwidth, clip, trim={0.1cm 0.1cm 0.1cm 0.1cm}]
    {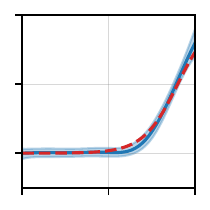}
    \caption*{Non-Gaussian likelihood-NUTS}
\end{subfigure}
\begin{subfigure}[t]{0.135\textwidth}
    \includegraphics[width=\textwidth, clip, trim={0.1cm 0.1cm 0.1cm 0.1cm}]  
    {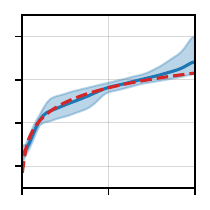}
    \includegraphics[width=\textwidth, clip, trim={0.1cm 0.1cm 0.1cm 0.1cm}]
    {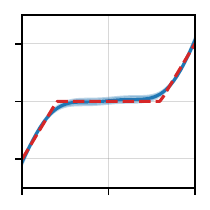}
    \includegraphics[width=\textwidth, clip, trim={0.1cm 0.1cm 0.1cm 0.1cm}]
    {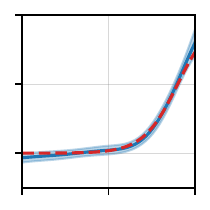}
    \caption*{RLRTO}
\end{subfigure}
    \caption{Unconstrained and constrained GP models for experiments with 1D functions. The red dash line is the ground truth. The blue line is the mean of the GP model with shaded area as the 95\% confidence interval. The orange circle markers are the virtual points. The black cross markers are the data points.}
    \label{fig:experiment-1d}
\end{figure}

%% file: examples_2d_mean.tex
\begin{figure}[H]
\centering
\begin{subfigure}[t]{\dimexpr0.135\textwidth+10pt\relax}
    \makebox[10pt]{\raisebox{20pt}{\rotatebox[origin=c]{90}{\footnotesize{2D-1}}}}%
    \includegraphics[width=\dimexpr\linewidth-10pt\relax]
    {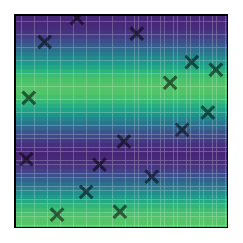}
    \makebox[10pt]{\raisebox{20pt}{\rotatebox[origin=c]{90}{\footnotesize{2D-2}}}}%
    \includegraphics[width=\dimexpr\linewidth-10pt\relax]
    {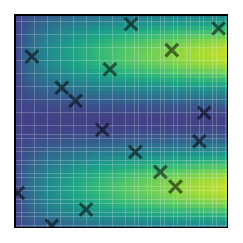}
    \makebox[10pt]{\raisebox{20pt}{\rotatebox[origin=c]{90}{\footnotesize{2D-3}}}}%
    \includegraphics[width=\dimexpr\linewidth-10pt\relax]
    {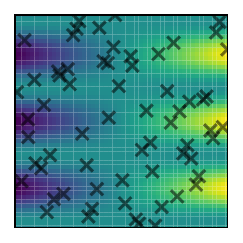}
    \caption*{Ground truth}
\end{subfigure}\hfill
\begin{subfigure}[t]{0.135\textwidth}
    \includegraphics[width=\textwidth]  
    {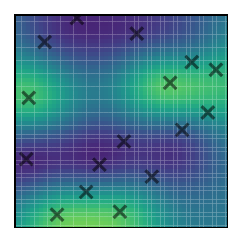}
    \includegraphics[width=\textwidth]
    {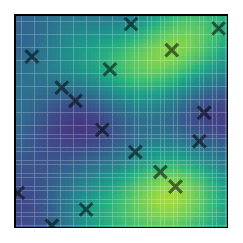}
    \includegraphics[width=\textwidth]
    {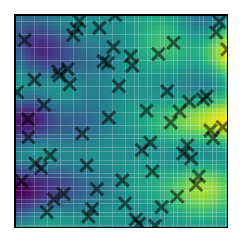}
    \caption*{Unconstrained}
\end{subfigure}\hfill
\begin{subfigure}[t]{0.135\textwidth}
    \includegraphics[width=\textwidth]  
    {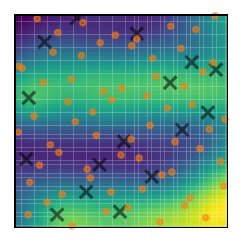}
    \includegraphics[width=\textwidth]
    {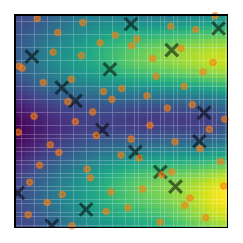}
    \includegraphics[width=\textwidth]
    {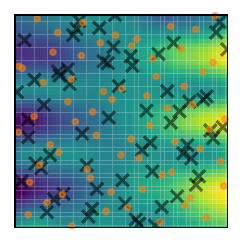}
    \caption*{Truncated prior-Gibbs}
\end{subfigure}\hfill
\begin{subfigure}[t]{0.135\textwidth}
    \includegraphics[width=\textwidth]  
    {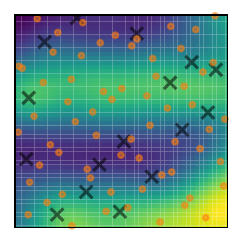}
    \includegraphics[width=\textwidth]
    {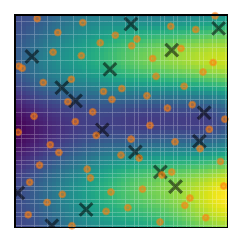}
    \includegraphics[width=\textwidth]
    {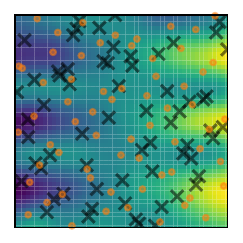}
    \caption*{Truncated prior-NUTS}
\end{subfigure}\hfill
\begin{subfigure}[t]{0.135\textwidth}
    \includegraphics[width=\textwidth]  
    {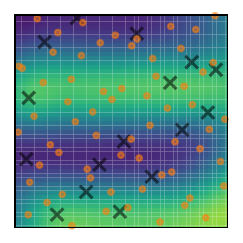}
    \includegraphics[width=\textwidth]
    {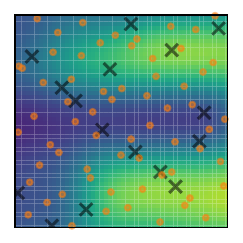}
    \includegraphics[width=\textwidth]
    {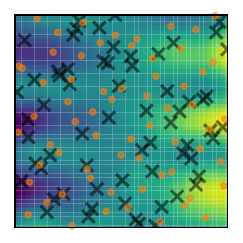}
    \caption*{Non-Gaussian likelihood-Gibbs}
\end{subfigure}\hfill
\begin{subfigure}[t]{0.135\textwidth}
    \includegraphics[width=\textwidth]  
    {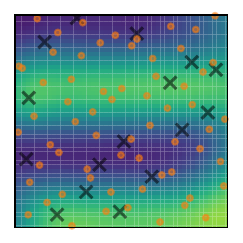}
    \includegraphics[width=\textwidth]
    {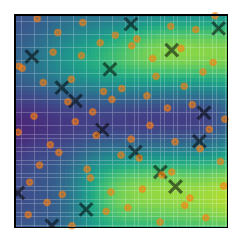}
    \includegraphics[width=\textwidth]
    {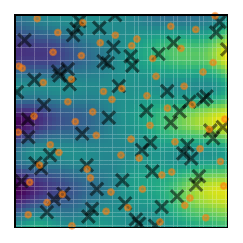}
    \caption*{Non-Gaussian likelihood-NUTS}
\end{subfigure}\hfill
\begin{subfigure}[t]{0.135\textwidth}
    \includegraphics[width=\textwidth]  
    {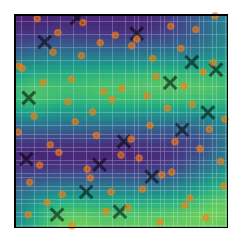}
    \includegraphics[width=\textwidth]
    {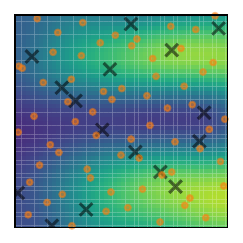}
    \includegraphics[width=\textwidth]
    {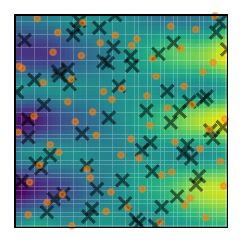}
    \caption*{RLRTO}
\end{subfigure}
    \caption{Means of the unconstrained and constrained GPs for the experimetns with 2D functions. Each row corresponds to a different experiment and each column corresponds to a different method except for the first column showing the ground truth. The colormap is the same for all subfigures in each row with yellow hues indicating high values. Black cross and orange circle markers indicate the locations of the data and virtual points, respectively.}
    \label{fig:experiment-2d-mean}
\end{figure}

%% file: examples_2d_CI.tex
\begin{figure}[H]
    \centering
\begin{subfigure}[t]{\dimexpr0.135\textwidth+10pt\relax}
    \makebox[10pt]{\raisebox{20pt}{\rotatebox[origin=c]{90}{\footnotesize{2D-1}}}}%
    \includegraphics[width=\dimexpr\linewidth-10pt\relax]
    {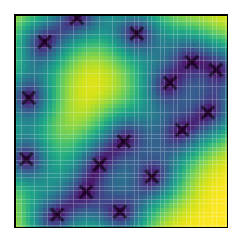}
    \makebox[10pt]{\raisebox{20pt}{\rotatebox[origin=c]{90}{\footnotesize{2D-2}}}}%
    \includegraphics[width=\dimexpr\linewidth-10pt\relax]
    {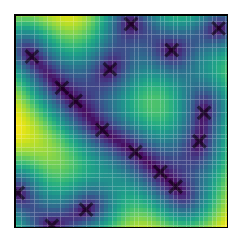}
    \makebox[10pt]{\raisebox{20pt}{\rotatebox[origin=c]{90}{\footnotesize{2D-3}}}}%
    \includegraphics[width=\dimexpr\linewidth-10pt\relax]
    {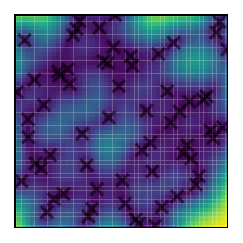}
    \caption*{Unconstrained}
\end{subfigure}
\begin{subfigure}[t]{0.135\textwidth}
    \includegraphics[width=\textwidth]  
    {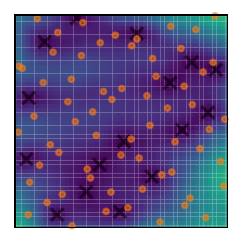}
    \includegraphics[width=\textwidth]
    {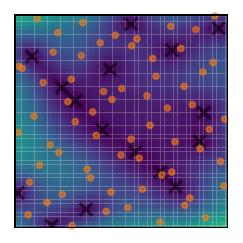}
    \includegraphics[width=\textwidth]
    {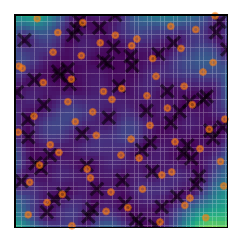}
    \caption*{Truncated prior-Gibbs}
\end{subfigure}
\begin{subfigure}[t]{0.135\textwidth}
    \includegraphics[width=\textwidth]  
    {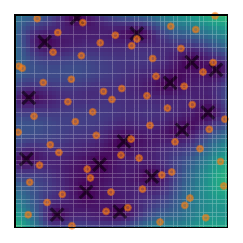}
    \includegraphics[width=\textwidth]
    {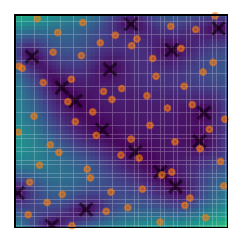}
    \includegraphics[width=\textwidth]
    {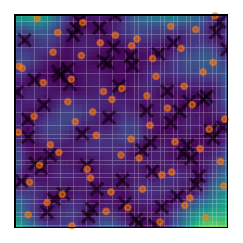}
    \caption*{Truncated prior-NUTS}
\end{subfigure}
\begin{subfigure}[t]{0.135\textwidth}
    \includegraphics[width=\textwidth]  
    {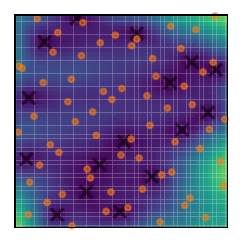}
    \includegraphics[width=\textwidth]
    {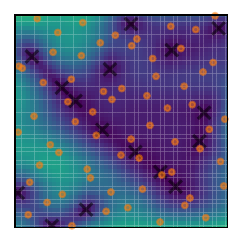}
    \includegraphics[width=\textwidth]
    {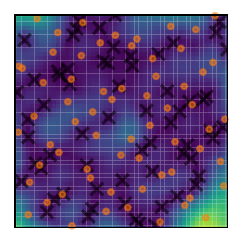}
    \caption*{Non-Gaussian likelihood-Gibbs}
\end{subfigure}
\begin{subfigure}[t]{0.135\textwidth}
    \includegraphics[width=\textwidth]  
    {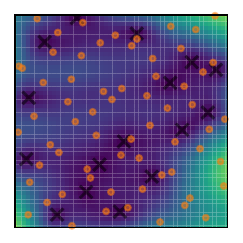}
    \includegraphics[width=\textwidth]
    {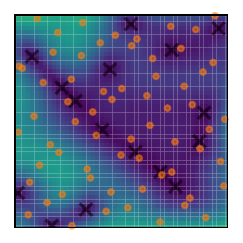}
    \includegraphics[width=\textwidth]
    {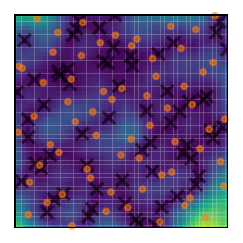}
    \caption*{Non-Gaussian likelihood-NUTS}
\end{subfigure}
\begin{subfigure}[t]{0.135\textwidth}
    \includegraphics[width=\textwidth]  
    {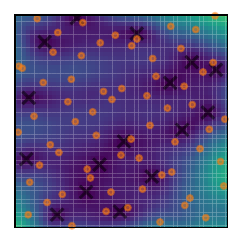}
    \includegraphics[width=\textwidth]
    {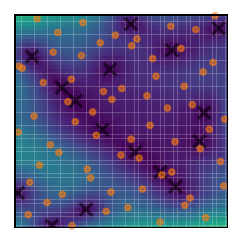}
    \includegraphics[width=\textwidth]
    {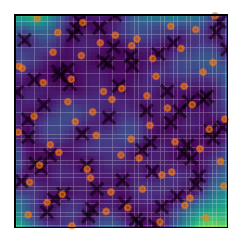}
    \caption*{RLRTO}
\end{subfigure}
    \caption{95\% credible intervals of the unconstrained and constrained GPs for the experiments with 2D functions. Each row corresponds to a different experiment and each column corresponds to a different method. The colormap is the same for all subfigures in each row with yellow hues indicating high values, and the colorbar is omitted as the exact magnitude of the values is not of interest. Black cross and orange circle markers indicate the locations of the data and virtual points, respectively.}
    \label{fig:experiment-2d-ci}
\end{figure}

%% file: applications.tex
\section{Applications}
\label{section:applications}
In this section, we will demonstrate the use of the proposed RLRTO method for building GP models for systems described by differential equations. We will consider two examples: the Susceptible-Infected-Removed (SIR) epidemic model and the convection diffusion equation \cite{langtangen2016scaling}. The same procedures as in Section~\ref{section:experiments} are followed to build the unconstrained and constrained GP models here. The truncated prior and non-Gaussian likelihood models are constructed solely using NUTS, as Gibbs sampling was found computationally inferior to NUTS in Section~\ref{section:experiments}.

\subsection{SIR epidemic model}
The spread of infectious diseases can be modeled by the SIR model, in which the population is divided into three compartments: susceptible (S), infected (I) and removed (R). Let $S(t)$, $I(t)$ and $R(t)$ be the number of individuals in each category at time $t$. We consider the dimensionless form of the SIR model, which is governed by the following system of ordinary differential equations (ODEs):
\begin{equation}
    \begin{aligned}
        \frac{dS}{dt} &= -R_0 S I, \\
        \frac{dI}{dt} &= R_0 S I - I, \\
        \frac{dR}{dt} &= I,
    \end{aligned}\label{eq:ode-sir}
\end{equation}
where $R_0$ is the basic reproduction number, which is a measure of the disease's ability to spread in a population. Here, we use the initial conditions $S(0) = 0.98$, $I(0) = 0.02$ and $R(0) = 0$. The solution of the SIR model is a three-dimensional function of time $t$ and the parameter $R_0$, i.e., $S(t, R_0)$, $I(t, R_0)$ and $R(t, R_0)$. It can be seen from \eqref{eq:ode-sir} that the $R(t, R_0)$ is monotonic is both $R_0$ and $t$. In this example, we will build a GP model to approximate the mapping from the parameter $R_0$ and time $t$ to the solution $R(t, R_0)$.

\begin{figure}[tb]
    \centering
    \begin{subfigure}{0.32\textwidth}
        \includegraphics[width=\textwidth]{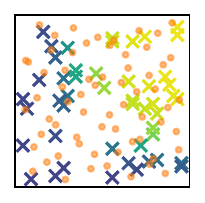}
        \caption{Data}
        \label{fig:sir-data}
    \end{subfigure}
    \hfill
    \begin{subfigure}{0.32\textwidth}
        \includegraphics[width=\textwidth]{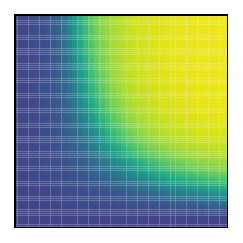}
        \caption{Ground truth}
        \label{fig:sir-ground-truth}
    \end{subfigure}
    \hfill
    \begin{subfigure}{0.32\textwidth}
        \includegraphics[width=\textwidth]{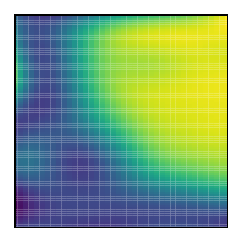}
        \caption{Unconstrained}
        \label{fig:sir-unconstrained}
    \end{subfigure}
    \\
    \begin{subfigure}{0.32\textwidth}
        \includegraphics[width=\textwidth]{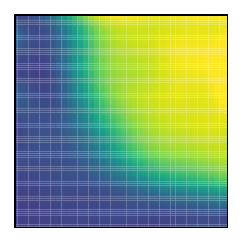}
        \caption{Truncated NUTS}
        \label{fig:sir-truncated-nuts}
    \end{subfigure}
    \hfill
    \begin{subfigure}{0.32\textwidth}
        \includegraphics[width=\textwidth]{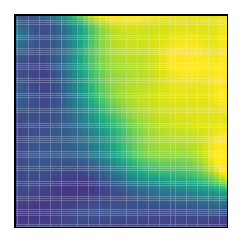}
        \caption{Nonlinear NUTS}
        \label{fig:sir-nonlinear-nuts}
    \end{subfigure}
    \hfill
    \begin{subfigure}{0.32\textwidth}
        \includegraphics[width=\textwidth]{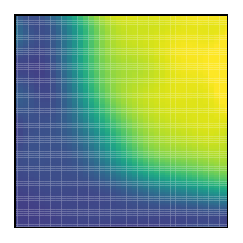}
        \caption{RLRTO}
    \end{subfigure}
    \caption{Means of the GPs in the SIR example.}
    \label{fig:sir}
\end{figure}

The data are generated by solving the ODE system \eqref{eq:ode-sir} with $R_0 \in [0.01, 5]$ and $t \in [0, 10]$ using the Runge-Kutta method of order 5,  with the implementation in \texttt{SciPy}. The solution is sampled at different values of $R_0$ and $t$ from a uniform distribution, resulting in the training data of 64 measurements, shown in Fig.~\ref{fig:sir-data} as cross markers. The rest of the data is used for testing and is shown as the groundtruth in Fig.~\ref{fig:sir-ground-truth}.

Using only the training points, the unconstrained GP model captures the general structure of the solution, as shown in Fig.~\ref{fig:sir-unconstrained}. However, the unconstrained GP model fails to preserve the solution's monotonicity, particularly evident in its inaccurate predictions along the domain's top and left boundaries. With 64 virtual points sampled from a Sobol sequence, the three constrained GP models improves the accuracy in terms of the MSE and reduces the 95\% CI width, with the RLRTO GP model achieving the best performance, as shown in Table~\ref{tab:sir-performance}. The RLRTO GP model also gives the lowest IAT and the highest ESS per second, indicating that it is the most efficient method among the three constrained GP models.

\begin{table}[tb]
    \centering
    \begin{tabular}{lcccc}
        \hline
        Method & MSE$/10^{-3}$ & 95\% CI width$/10^{-2}$ & IAT & ESS per second \\
        \hline
        Unconstrained  & 9.06  & 17.8  & NA       & NA \\
        Truncated prior-NUTS & 1.23 & 4.07 & 6.84 & 0.928 \\
        Non-Gaussian likelihood-NUTS & 3.07 & 6.03 & 134 & $8.26\cdot10^{-4}$ \\
        RLRTO            & \textbf{0.986} & \textbf{3.83} & \textbf{1.09} & \textbf{8.16} \\
        \hline
    \end{tabular}
    \caption{Performance of the unconstrained and constrained GP models in the SIR example. Bold values indicate the best metric performance.}
    \label{tab:sir-performance}
\end{table}

\subsection{Convection diffusion equation}
Here we consider building GP models to approximate the mapping from the parameters of a PDE system and its solution. Specifically, we consider the transient convection diffusion equation in 1D, given by
\begin{equation}
    \frac{\partial u}{\partial t} + b \frac{\partial u}{\partial x}- \alpha \frac{\partial^2 u}{\partial x^2} = 0, \quad x \in [0, 1], t > 0,\label{eq:conveciton-diffusion}
\end{equation}
where $u(x, t)$ is the temperature at position $x$ and time $t$, and $\alpha$ is the thermal diffusivity constant. We consider a constant velocity $b<0$, which means the convective flow is in the negative direction of the 1D domain. The diffusivity $\alpha$ is fixed to be 0.1. The initial condition is $u(x, 0) = 0$ and we set a Dirichlet boundary at the upper boundary $u(1, t) = 1$ for all $t > 0$ and omit the lower boundary condition at $x=0$. Our goal is to approximate the solution $u$ at given velocity $b$, time $t$, and position $x$, so this is a three-dimensional problem with the input space $\mathbb{R}^3$ and the output space $\mathbb{R}$. Based on the physical setting of the problem, it is reasonable to expect that the solution is monotonic with respect to each of the input variables: $b$, $x$, and $t$.

\begin{figure}[tb]
    \centering
    \begin{subfigure}{0.32\textwidth}
        \includegraphics[width=\textwidth, clip, trim={0.5cm 0.5cm 0.0cm 0.5cm}]{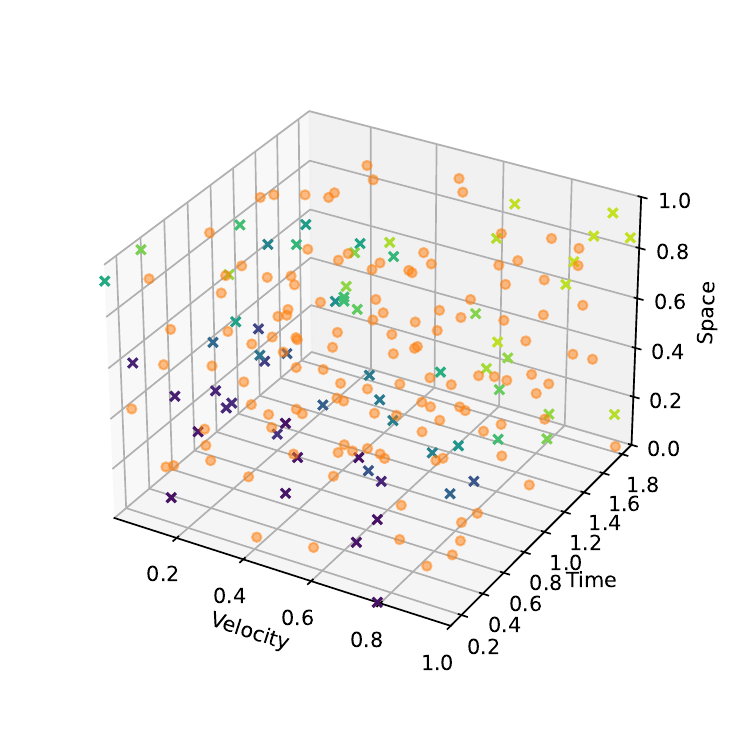}
        \caption{Data}
        \label{fig:convection-diffusion-data}
    \end{subfigure}
    \hfill
    \begin{subfigure}{0.32\textwidth}
        \includegraphics[width=\textwidth, clip, trim={0.5cm 0.5cm 0.0cm 1.5cm}]{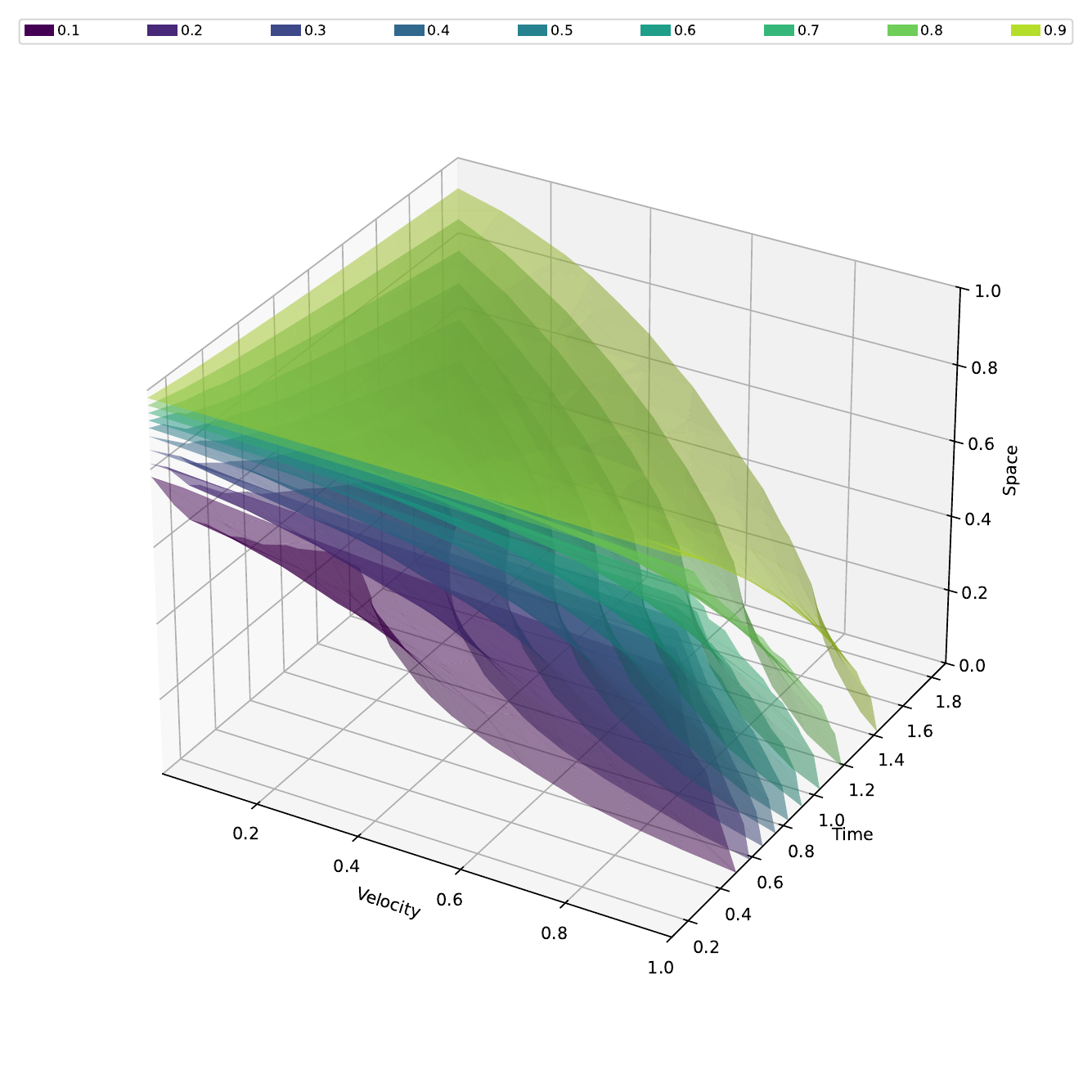}
        \caption{Ground truth}
        \label{fig:convection-diffusion-ground-truth}
    \end{subfigure}
    \hfill
    \begin{subfigure}{0.32\textwidth}
        \includegraphics[width=\textwidth, clip, trim={0.5cm 0.5cm 0.0cm 1.5cm}]{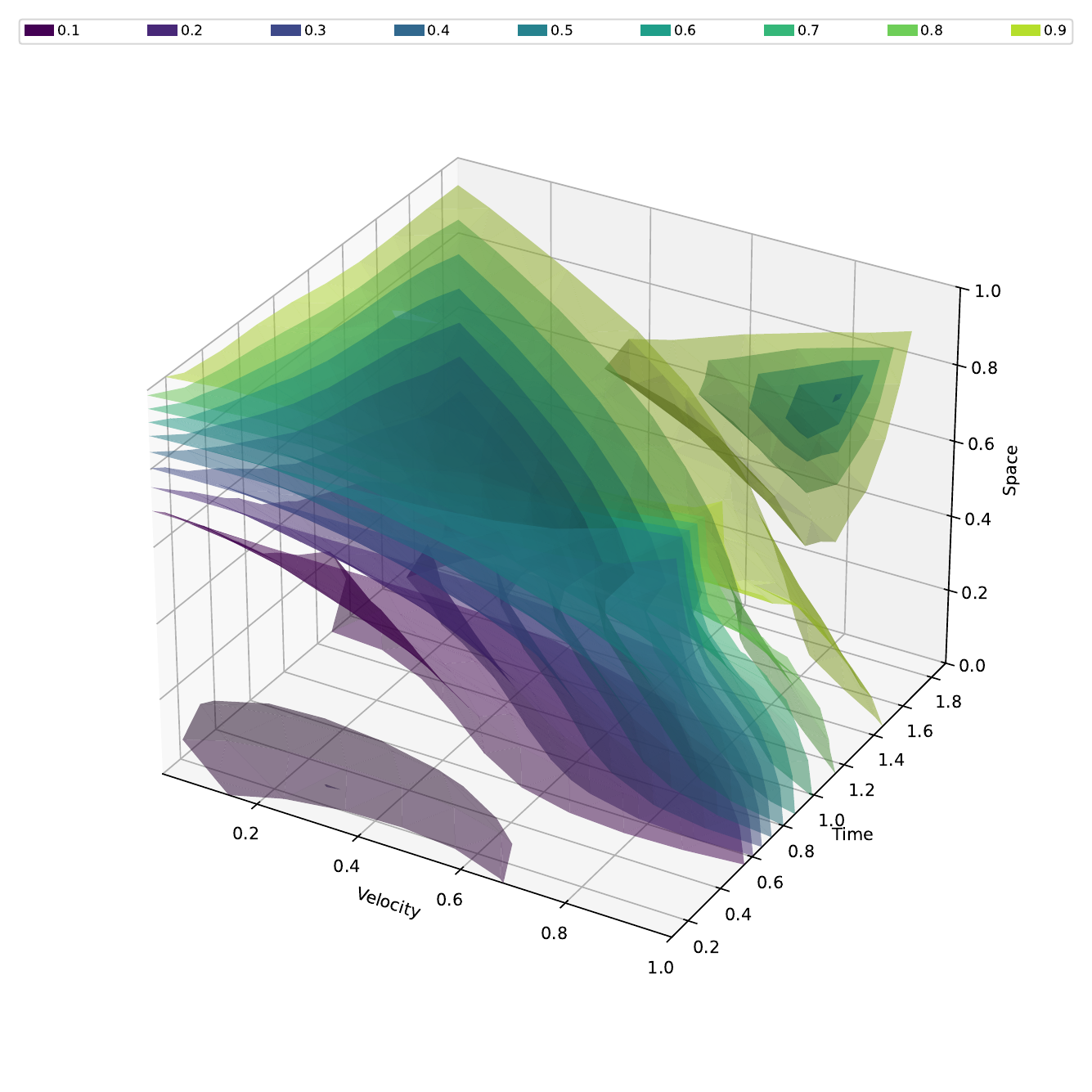}
        \caption{Unconstrained}
    \end{subfigure}
    \\
    \begin{subfigure}{0.32\textwidth}
        \includegraphics[width=\textwidth, clip, trim={0.5cm 0.5cm 0.0cm 1.5cm}]{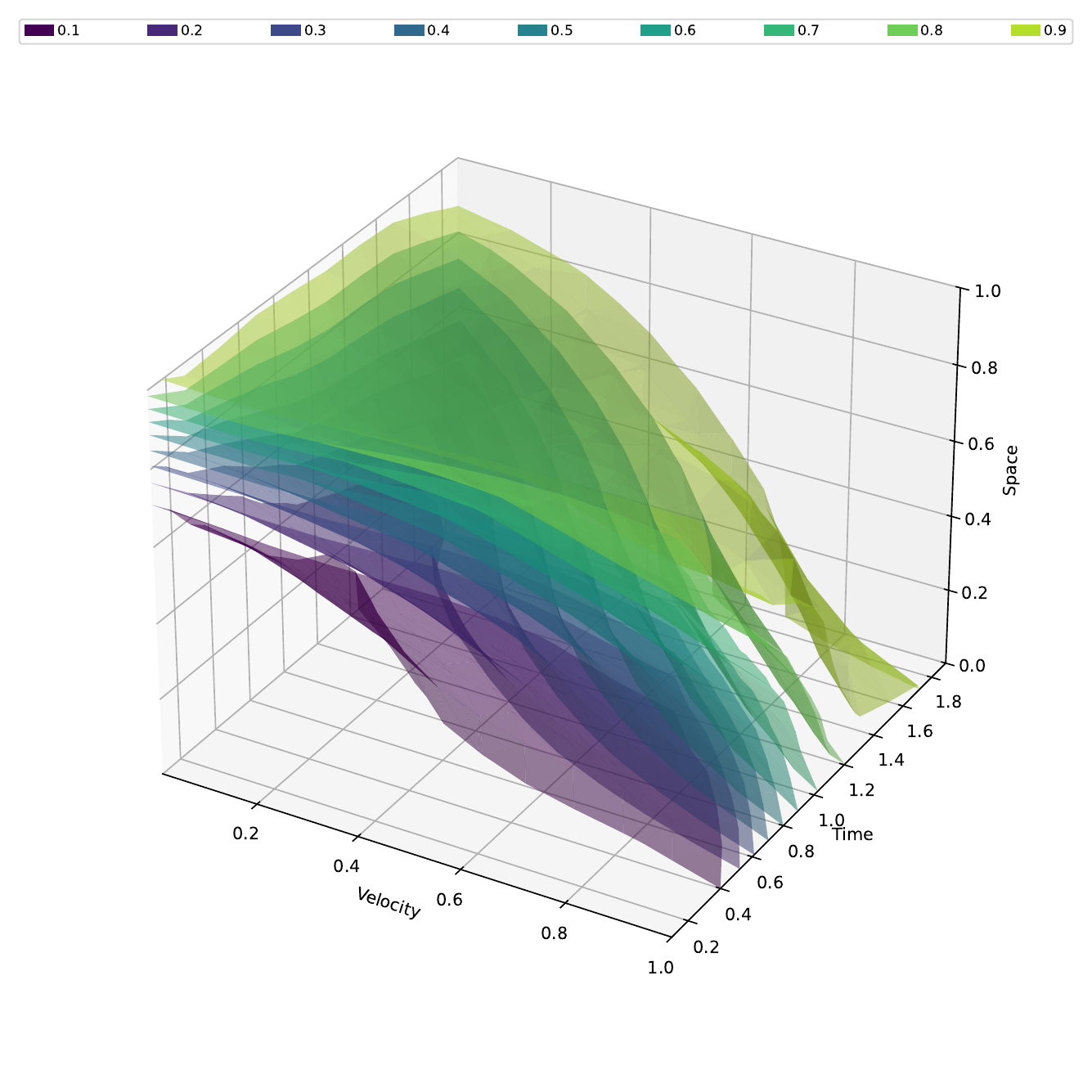}
        \caption{Truncated prior-NUTS}
    \end{subfigure}
    \hfill
    \begin{subfigure}{0.32\textwidth}
        \includegraphics[width=\textwidth, clip, trim={0.5cm 0.5cm 0.0cm 1.5cm}]{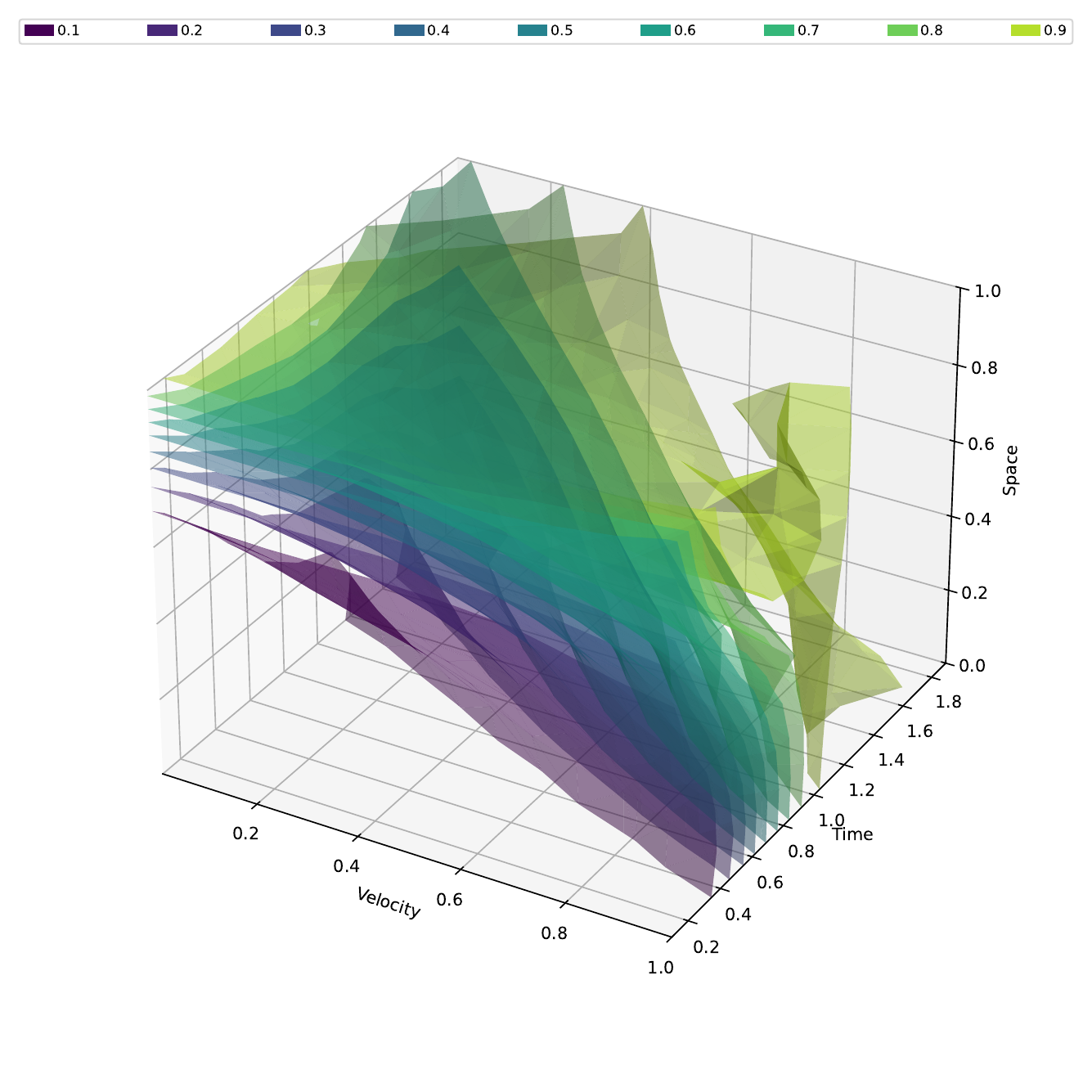}
        \caption{Non-Gaussian likelihood-NUTS}
        \label{fig:convection-diffusion-NUTS}
    \end{subfigure}
    \hfill
    \begin{subfigure}{0.32\textwidth}
        \includegraphics[width=\textwidth, clip, trim={0.5cm 0.5cm 0.0cm 1.5cm}]{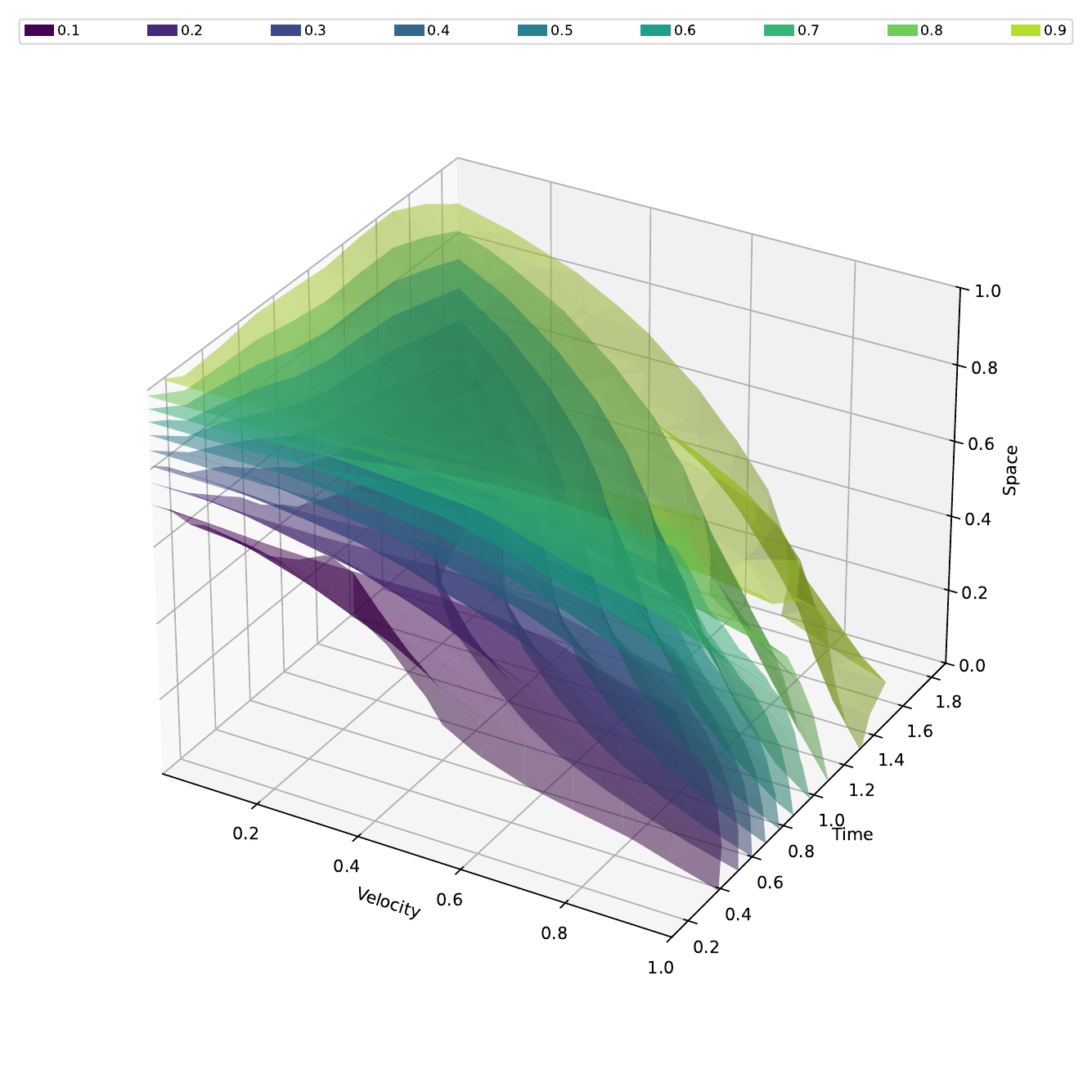}
        \caption{RLRTO}
    \end{subfigure}
    \caption{Mean isosurfaces of the GPs in the convection-diffusion example.}
    \label{fig:convection-diffusion}
\end{figure}

The finite element method (FEM) is used to discretize \eqref{eq:conveciton-diffusion} in space with 64 elements and a backward Euler scheme with a time step of 0.01 to integrate in time until $T=1.5$. The solution is computed at different velocities $b \in [-1, 0]$. The solution is sampled at 64 different velocities, time points, and positions from a uniform distribution, resulting in the training data of 64 measurements, shown in Fig.~\ref{fig:convection-diffusion-data} as cross markers. The remaining data are used for testing, an isosurface of which is shown in Fig.~\ref{fig:convection-diffusion-ground-truth}.

128 points from the Sobol sequence are used as the virtual points to enforce monotonicity, as indicated by orange circles in Fig.~\ref{fig:convection-diffusion-data}. As shown in Fig.~\ref{fig:convection-diffusion}, the unconstrained GP is not able to capture the monotonicity of the solution, while the RLRTO GP is able to capture the monotonicity and provides a better approximation of the solution. The improvement is also reflected in the MSE, which is $3\cdot10^{-3}$ for the unconstrained GP and $9.96\cdot 10^{-4}$ for the RLRTO GP. Across all the models, the truncated prior and the RLRTO models show comparable performance in terms of MSE and 95\% CI width, while
the non-Gaussian likelihood model performs noticeably worse, as detailed in Table~\ref{tab:convection-diffusion-performance}. The RLRTO model shows the best efficiency performance with the lowest IAT and the highest ESS per second.

\begin{table}[tb]
    \centering
    \begin{tabular}{lcccc}
        \hline
        Method & MSE$/10^{-3}$ & 95\% CI width$/10^{-2}$ & IAT & ESS per second \\
        \hline
        Unconstrained  & 3.00 & 11.0 & NA       & NA \\
        Truncated prior-NUTS & \textbf{0.884} & \textbf{4.00} & 6.91 & 0.830 \\
        Non-Gaussian likelihood-NUTS & 2.05 & 5.58 & 10.7 & 0.198 \\
        RLRTO            & 0.996 & 4.12 & \textbf{1.27} & \textbf{1.38} \\
        \hline
    \end{tabular}
    \caption{Performance of the unconstrained and constrained GP models in the convection diffusion example. Bold values indicate the best metric performance.}
    \label{tab:convection-diffusion-performance}
\end{table}

%% file: conclusions.tex
\section{Conclusions and future work}
\label{section:conclusion}

In this work, we have proposed a novel approach to the problem of constraining GPs with monotonicity constraints. Our method leverages the RLRTO framework for efficient sampling of a constrained posterior distribution of the derivative GPs at virtual points. We have demonstrated that in many cases our approach produces at least as good results as two existing virtual point-based methods in the literature, i.e., the truncated prior and non-Gaussian likelihood methods, whilst being more efficient, especially for the more computationally demanding larger problems with an increasing number of virtual points. The proposed method is particularly useful for applications where monotonicity constraints are important, such as in optimization and decision-making problems.

Meanwhile, we have also proposed to enhance the two existing virtual point-based methods by replacing the computationally intensive Gibbs sampler with the state-of-the-art NUTS. Our experiments demonstrate that NUTS achieves better computational performance than Gibbs sampling, particularly for problems with many virtual points.

An important future direction is to apply and compare the proposed methods beyond the scope of monotonicity constraints for Gaussian processes. In particular, incorporating constraints other than monotonicity should be relatively straightforward to implement when projection mappings onto the constraint set are available \cite{everink2023sparse}. Furthermore, these approaches are not limited to constrained GPs, but can be applied to many other Bayesian inference problems.